\documentclass[3p,times]{elsarticle}
\usepackage{ecrc}
\usepackage{algorithm}
\usepackage{algorithmic}

\usepackage{hyperref}       
\usepackage{url}            
\usepackage{booktabs}       
\usepackage{amsfonts}       
\usepackage{nicefrac}       
\usepackage{microtype}      
\usepackage{xcolor}         
\usepackage{bm,bbm}
\usepackage{amsmath}
\usepackage{amsthm}
\usepackage{graphicx}
\usepackage{multirow}
\usepackage{subcaption,booktabs}

\volume{00}

\firstpage{1}

\journalname{Journal of XXXXXX}

\runauth{}

\jid{procs}

\jnltitlelogo{Journal of XXXXXX}

\usepackage{amssymb}

\usepackage[figuresright]{rotating}
\usepackage{setspace}

\begin{document}

\begin{frontmatter}

\title{FV-MgNet: Fully Connected V-cycle MgNet for Interpretable Time Series Forecasting}


\author[1]{Jianqing Zhu}
\ead{jqzhu@emails.bjut.edu.cn}

\author[2]{Juncai He}
\ead{juncai.he@kaust.edu.sa}
\ead[url]{https://juncaihe.github.io}

\author[3]{Lian Zhang}
\ead{zhanglian@sribd.cn}

\author[2,4]{Jinchao Xu\corref{cor1}}
\ead{xu@multigrid.com}
\ead[url]{http://www.personal.psu.edu/jxx1/}
\cortext[cor1]{Corresponding author}

\address[1]{{Faculty of Sciences}, {Beijing University of Technology}, {{Chaoyang}, {Beijing}, {100124}, {Beijing}, {People’s Republic of China}}}

\address[2]{{Computer, Electrical and Mathematical Science and Engineering Division}, {King Abdullah University of Science and Technology (KAUST)}, {{Thuwal}, {23955-6900}, {Saudi Arabia}}}

\address[3]{{Shenzhen International Center for Industrial and Applied Mathematics, Shenzhen Research Institute of Big Data}, {2001 Longxiang Boulevard, Longgang District}, {{Shenzhen}, {518172}, {Guangdong},{People’s Republic of China}}}

\address[4]{Department of Mathematics, Pennsylvania State University, University Park, PA 16802, USA}

\begin{abstract}
By investigating iterative methods for a constrained linear model, we propose a new class of fully connected V-cycle MgNet for long-term time series forecasting, which is one of the most difficult tasks in forecasting.
MgNet is a CNN model that was proposed for image classification based on the multigrid (MG) methods for solving discretized partial differential equations (PDEs). We replace the convolutional operations with fully connected operations in the existing MgNet and then apply them to forecasting problems. Motivated by the V-cycle structure in MG, we further propose the FV-MgNet, a V-cycle version of the fully connected MgNet, to extract features hierarchically. 
By evaluating the performance of FV-MgNet on popular data sets and comparing it with state-of-the-art models, we show that the FV-MgNet achieves better results with less memory usage and faster inference speed. In addition, we develop ablation experiments to demonstrate that the structure of FV-MgNet is the best choice among the many variants.
\end{abstract}

\end{frontmatter}


\section{Introduction}\label{introduction}
Long-term time series forecasting has played an important role in numerous applications across an array of sectors, including retail~(\cite{bose2017probabilistic,courty1999timing}), healthcare~(\cite{lim2018forecasting,zhang2018multi}), and engineering~(\cite{zhang2019deep,gonzalez2019methodology}). Various deep learning models have been developed for time series forecasting, among which recurrent neural networks (RNN) are probably the most extensively studied~(\cite{connor1994recurrent,hewamalage2021recurrent}). Recently, following its success in natural language processing (NLP) and computer vision (CV) research~(\cite{attention_is_all_you_need,Bert/NAACL/Jacob,Transformers-for-image-at-scale/iclr/DosovitskiyB0WZ21,DBLP:Global-filter-FNO-in-cv}), the Transformer model has become one of the most popular research directions in time series forecasting~(\cite{Log-transformer-shiyang-2019,haoyietal-informer-2021,Autoformer}). Efficient self-attention has been introduced, and many  structures have been proposed to handle time-series forecasting tasks, such as decomposing the sequence~(\cite{Autoformer}), designing a special linear attention structure~(\cite{Log-transformer-shiyang-2019,haoyietal-informer-2021}), and applying Fourier transform ~(\cite{Autoformer}) or wavelet transform~(\cite{zhou2022fedformer}) to the self-attention structure. However, the Transformer model has a high computational cost, and many experiments have found that forecasting results degenerate as the length of the input sequence increases ~(\cite{wen2022transformers}).

\begin{figure}[htbp]
	\centering
	\includegraphics[width=\textwidth]{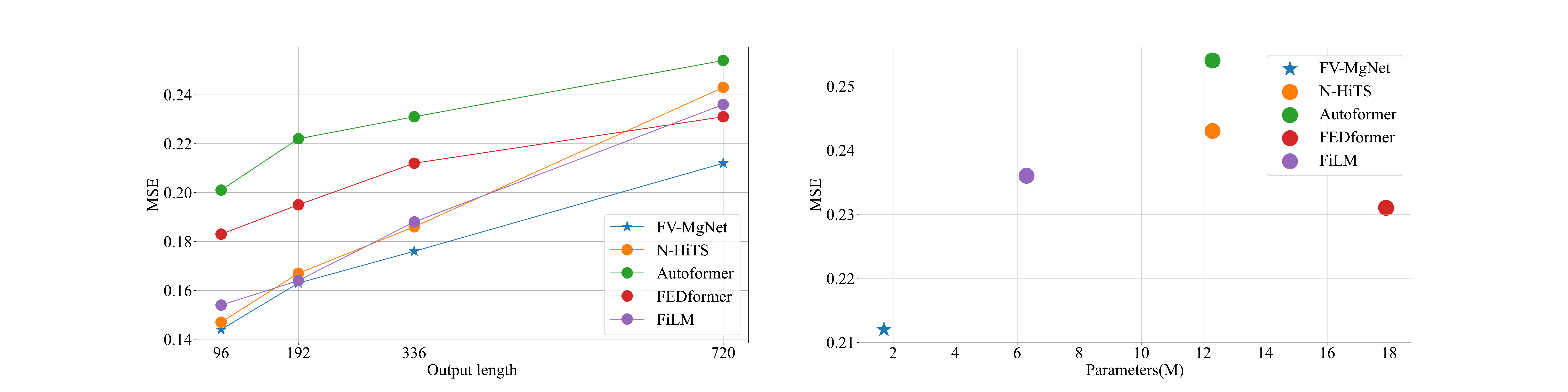}
	\caption{MSE on the Electricity multivariate data set. Left: MSE
		as output length. Right: MSE as parameters, output length is fixed to 720. A lower MSE indicates a better prediction. }
	\label{fig:mse-param}
\end{figure}

\begin{figure}[htbp]
	\centering
	\includegraphics[width=\textwidth]{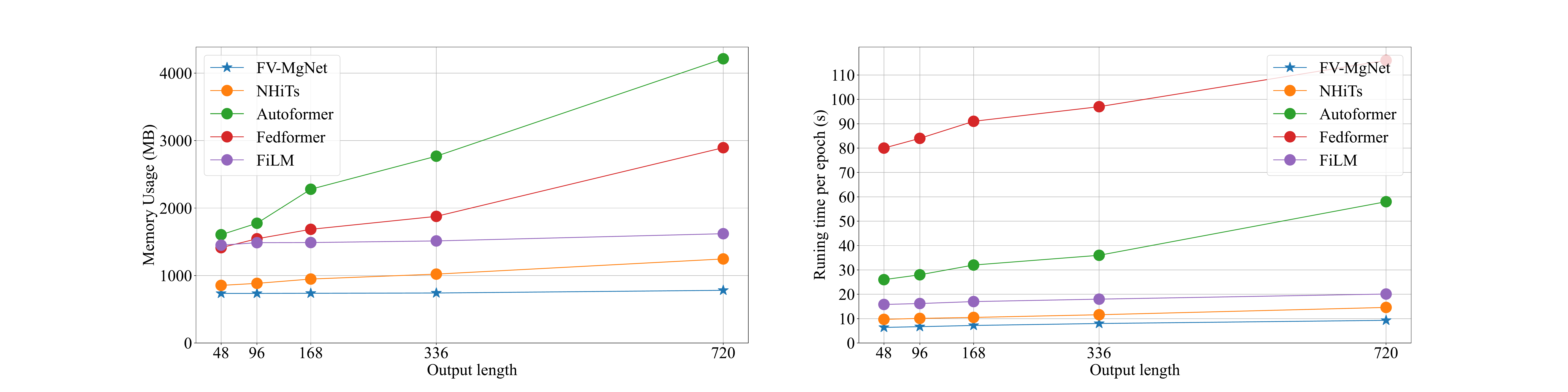}
	\caption{Computational efficiency benchmark on the Electricity multivariate data set, batch size is 16. Time usage is measured for one epoch training time. Left: Memory usage
		as output length, input length is fixed to 96. Right: Epoch time usage as output length, input length is fixed to 96. }
	\label{fig:memory-time}
\end{figure}

By investigating iterative methods for a constrained linear model, we propose a new class of fully connected V-cycle MgNet (FV-MgNet) for long-term time series forecasting. This is the first time an MgNet type model has successfully been applied to this field. One of the main technologies  FV-MgNet is used for is to replace the convolutional operations with fully connected operations in the existing MgNet~(\cite{he2019mgnet,he2021interpretive,wang2022cnns}), which is a CNN model proposed for image classification based on the multigrid (MG) methods~(\cite{xu1992iterative,xu2002method,xu2017algebraic}) for solving discretized partial differential equations (PDEs).  
The other technology, that we propose, is a V-cycle version of the fully connected MgNet to extract features hierarchically, motivated by the V-cycle structure in MG, where different frequency information can be captured through the size transformation between grids. 
By evaluating the performance of the FV-MgNet on popular data sets and comparing it with state-of-the-art models, we present that the FV-MgNet achieves better results with much less memory usage and faster inference speed (as shown in Figure\ref{fig:mse-param}~ and Figure~\ref{fig:memory-time}). We also demonstrate that numerical results do not degenerate as the length of the input increases (as shown in Table~\ref{tab:prolong-input-length}). 

This paper is organized as follows. In section \ref{relatedwork}, we review some of the related work. In section \ref{models}, we first consider the time series forecasting problem as a constrained linear model and solve it using the multigrid method, an iterative method, and then we introduce  FV-MgNet models for forecasting problems. Numerical experiments and a Conclusion are given in Sections \ref{experiments} and \ref{conclution}, respectively.

\section{Related Work}\label{relatedwork}
Recurrent neural networks are widely used in time series prediction problems. LSTM~(\cite{graves2012long}) and GRU ~(\cite{chung2014empirical}) can alleviate the inherent problems such as gradient disappearance and gradient explosion through a gating mechanism. Many works based on LSTM and GRU have proposed numerous time series prediction models (e.g. ~(\cite{wen2017multi,salinas2020deepar,guo2019exploring,fan2019multi}). However, the single-step prediction method cannot accurately capture long-term time series features, and it is difficult to parallelize. More recently, a convolutional neural network has been applied to this task and a temporal convolutional network (TCN) was proposed~(\cite{sen2019think}). Although this method is   efficient,  there is also a prediction bottleneck in the long-term time series prediction problem. With the innovation of Transformers in NLP and CV~(\cite{attention_is_all_you_need,Bert/NAACL/Jacob,Transformers-for-image-at-scale/iclr/DosovitskiyB0WZ21}), Transformer-based models have also been proposed, and perform well, in time series forecasting~(\cite{Log-transformer-shiyang-2019,haoyietal-informer-2021,Autoformer}). Transformer-based models excel in modeling long-term dependencies for sequential data. Furthermore, many efficient Transformers are designed in time series forecasting to overcome the quadratic computation complexity of the original Transformer without performance degradation. LogTrans~(\cite{Log-transformer-shiyang-2019}) adopts log-sparse attention and achieves $N\log^2 N$ complexity. Reformer~(\cite{DBLP:conf/iclr/KitaevKL20-reformer}) introduces a local-sensitive hashing that reduces the complexity to $N\log N$. Informer~(\cite{haoyietal-informer-2021}) uses a KL-divergence-based method to select top-k in the attention matrix and costs $N\log N$ in complexity. Most recently, Autoformer~(\cite{Autoformer}) introduced an auto-correlation block in place of canonical attention to perform sub-series level attention, which achieves $N\log N$ complexity with the help of Fast Fourier transform (FFT) and top-k selection in the auto-correlation matrix. Fedformer~(\cite{zhou2022fedformer}) designs frequency domain attention by applying Fourier or wavelet transform. ETSformer~(\cite{woo2022etsformer}) proposes exponential smoothing attention (ESA) and frequency attention (FA), which improves accuracy and efficiency.

Recently, many non-Transformer models have also achieved good results in forecasting problems. FiLM~(\cite{zhou2022film}) applies Legendre Polynomial projections, Fourier projections, and low-rank approximations in time series forecasting. The NBEATS~(\cite{oreshkin2019n}) model was proposed to design a double residual structure with fully connected layers, which has achieved good results in univariate prediction. N-HiTS~(\cite{challu2022n}) adds a multi-scale structure to NBEATs, so that it can better capture the information from different periods in the sequence and achieve improved results in a multivariate model. In NBEATS and N-HiTS, we observe fully connected layers; multi-scale structures are important to help improve model performance. MgNet~(\cite{he2019mgnet, wang2022cnns}) is a CNN model proposed for image classification based on the multigrid, which has been a successful method for solving PDEs. MgNet is explained by using multigrid ideas to solve a constrained linear model~(\cite{he2021interpretive}) so that  it has a well-designed multiscale structure. We investigate the fully connected layers and  multiscale structure under the MgNet framework to solve forecasting problems.

\section{Methodology}\label{models}

In this section, we present our proposed model, FV-MgNet, for time series forecasting using the constrained linear model and a multigrid structure; a detailed description of FV-MgNet is given in Algorithm~\ref{alg:mgnet}.
Considering the data-label pair as $(f,y)$ where $f\in \mathbb R^I$ is the input time series and $y\in \mathbb R^O$ is the predicted time series, we propose the following feature extraction and interpolation scheme as an interpretable model to build the mapping from $f$ to $y$:
\begin{equation}
	\text{Input:}~f \xrightarrow[\text{constrained linear model: }Au=f]{\text{feature extraction}} \text{Feature:}~u \xrightarrow[\text{neural network approximation}]{\text{feature interpolation: }\mathcal I} \text{Output:}~y.
\end{equation}
To summarise, we have
\begin{equation}
	y = \mathcal I(u(f)).
\end{equation}
Here, we denote $u\in \mathbb R^{I'}$ as the feature vector. In general, the information contained in the feature (essential dimension) should be less than or equal to the information from the data (embedded dimension), such as in, for example, manifold learning~(\cite{tenenbaum2000global,roweis2000nonlinear,donoho2003hessian}); precisely, $I'\le I$. To extract more intrinsic features but retain as much information as possible, we take $I'=I$ while adopting a V-cycle structure (similar to U-Net architecture) for feature extraction.
First, we propose to apply the constrained liner model~(\cite{he2021interpretive}), which was originally designed for vision tasks for feature extraction. 
Then, we propose to apply the fully connected operator as the underlying linear model instead of the convolutional operator in image processing.
Finally, we introduce the multigrid structure to construct features in multi-resolution and reduce the computational complexity.
After obtaining the feature vector $u$, we propose to approximate $\mathcal I(u)$ by using a one-hidden layer ReLU neural network that can be expressed as
\begin{equation}
	\label{eq:fI}
	\mathcal I(u) \approx W^2\sigma\left(W^1u\right).
\end{equation}

The proposed model uses the multigrid method as the main framework, performs residual correction on the input of different resolutions, integrates the information on other grids through the extension operator, and then adopts a one-hidden layer ReLU neural network approximation for interpolation to obtain the long horizon accurate predictions. In addition, compared to models based on transformer architecture, our model is implemented with fully connected layers, which has a more elegant structure with higher computational efficiency and lower memory overhead.

\subsection{Constrained linear  model: An interpretable model for feature extraction}
The constrained linear  model~(\cite{he2021interpretive}) was originally proposed for vision tasks; in this work, we demonstrate that it can also be applied as an interpretable model for feature extraction in time series problems. 
For any input time series $f\in \mathbb R^{I}$ in data space, the constrained linear model is defined as the following data-feature mapping
\begin{equation}\label{Auf}
	A u = f,
\end{equation}
where $u \in \mathbb R^{I'}$ is the underlying feature vector such that
\begin{equation}
	\label{positive-u}
	[u]_i \ge 0, \quad i=1: I'.
\end{equation}
Here, we assume that $A: \mathbb R^{I} \mapsto \mathbb R^{I'}$ is a linear (affine) operator from the feature space to the data space. It is crucial to solving the linear \eqref{Auf} with a nonlinear constraint~\eqref{positive-u}. Typically, the iterative method is an important part of solving algebraic systems. For example, given an algebraic system $Au=f$, the solution can be obtained through  the residual correction method
\begin{equation}
	u^{i+1} = u^{i}+B^i(f^{i}-Au^{i}), \quad i=1:\nu,
\end{equation}
where $B^i: \mathbb R^{I'} \mapsto \mathbb R^{I}$ is an approximation of the inverse of the data-feature mapping $A$.
Recalling the definition of the activation function $\sigma(x)={\rm ReLU}(x) := \max\{0,x\}$, the above iterative process can be naturally modified to
preserve the constraint \eqref{positive-u}: 
\begin{equation}
	\label{iterative}
	u^{i} = u^{i-1} + \sigma \circ B^i \sigma \circ  (f -  A u^{i-1}), \quad i=1:\nu.
\end{equation}
Here, $\sigma\circ(\cdot)$ denotes the element-wise applications of the nonlinear activation function $\sigma$.
We remark that $\sigma(x) = {\rm ReLU}(x)$ is not the unique choice to preserve the constraint~\eqref{positive-u}. For example, the Hat function~\cite{wang2022cnns} and the combination of ${\rm ReLU}^k(x)$~\cite{chen2022power} may also be used for this purpose.
As demonstrated in~(\cite{he2021interpretive}), the above iteration scheme can degenerate to the basic block of ResNet~(\cite{he2016deep}) if $r =f-Au$ is chosen as the feature. However, theoretical arguments and numerical results in~(\cite{he2021interpretive}) indicate that $u$ should be a more appropriate choice as the feature.
In this work, we also take \eqref{iterative} as the basic block in our model for feature extraction. 

\subsection{Underlying system A: Dense or sparse?}
\begin{figure}[h]
	\centering
	\begin{minipage}[t]{0.32\textwidth}
		\centering
		\includegraphics[width=\linewidth]{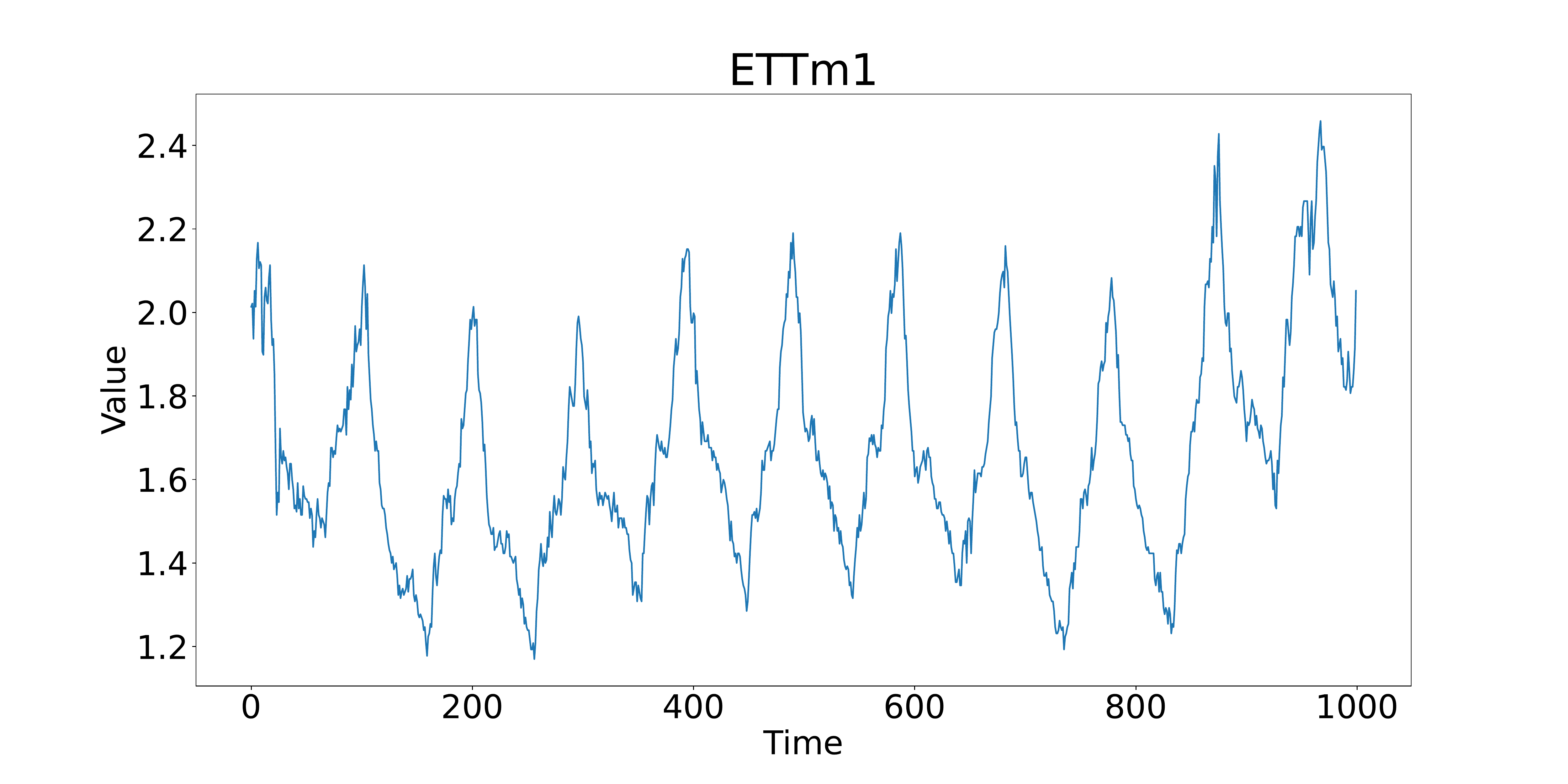}
	\end{minipage}
	\begin{minipage}[t]{0.32\textwidth}
		\centering
		\includegraphics[width=\linewidth]{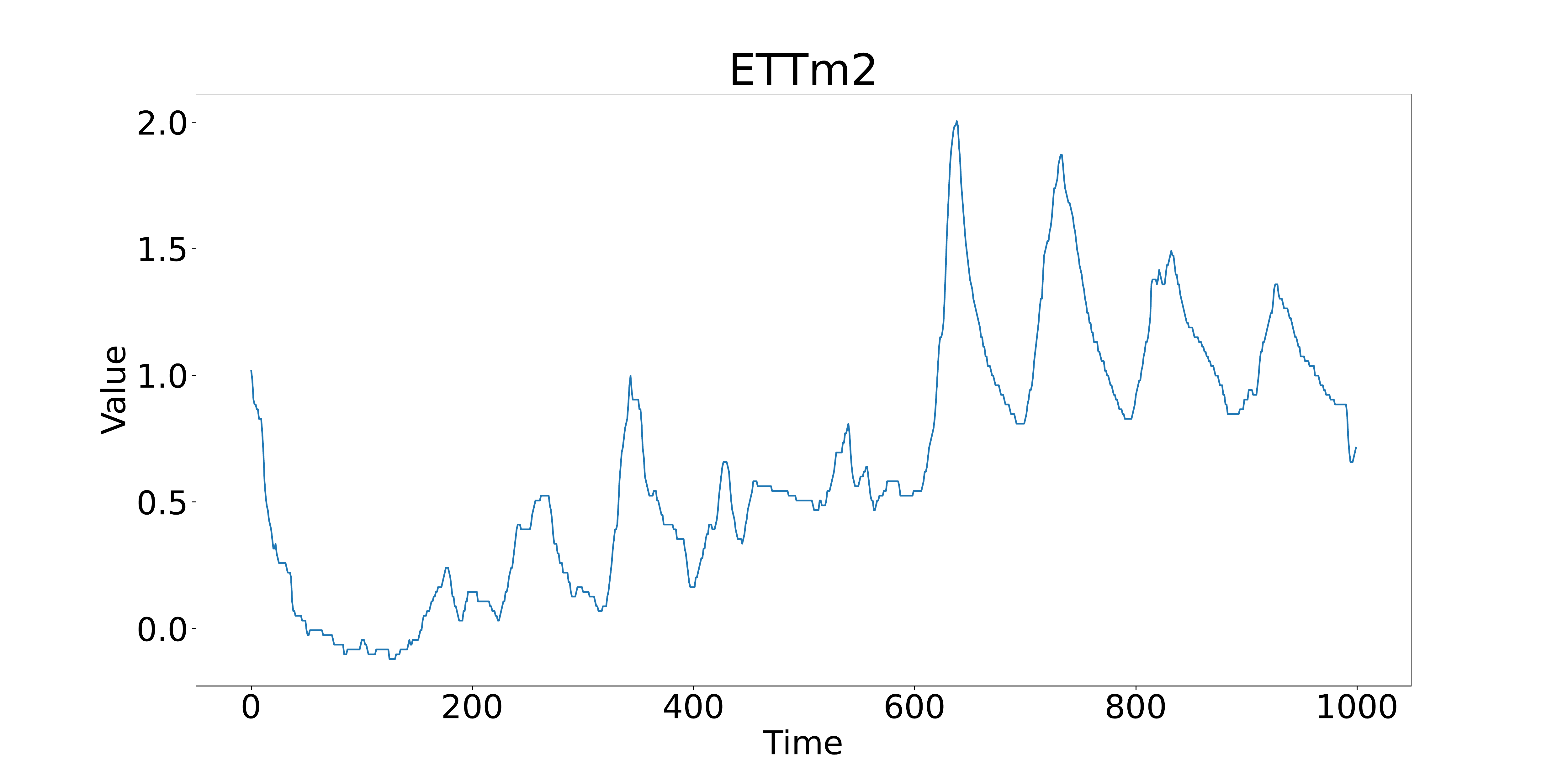}
	\end{minipage}
	\begin{minipage}[t]{0.32\textwidth}
		\centering
		\includegraphics[width=\linewidth]{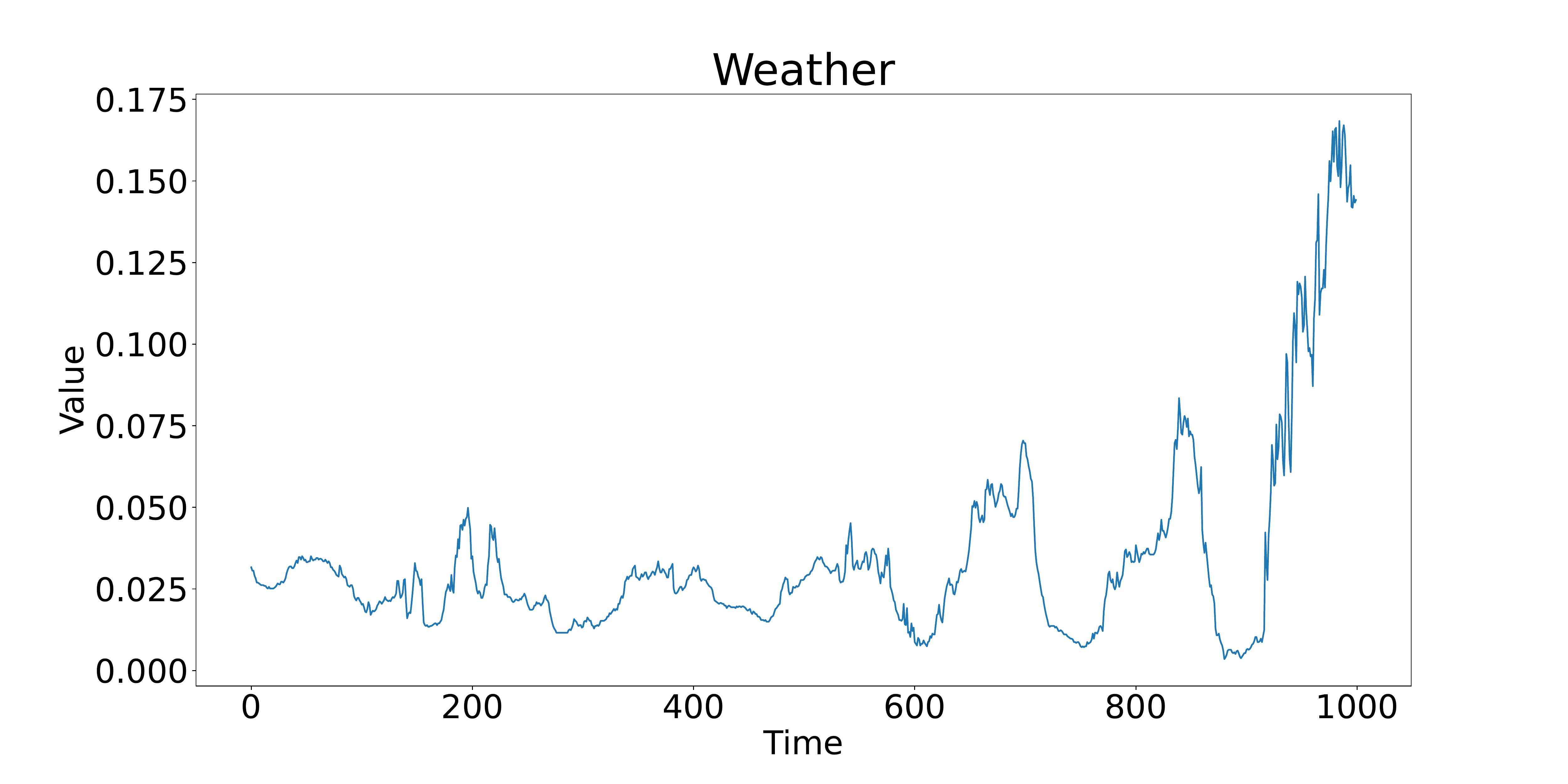}
	\end{minipage}
	\caption{Some typical long-term time series from different data sets.}
	\label{fig:dataset-freq}
\end{figure}

In vision tasks, $A$ and $B^i$ are generally chosen as learnable multi-channel convolution operators. For example, the basic block for image classification in MgNet~(\cite{he2019mgnet}) is:
$$u^{i} = u^{i-1} + \sigma \circ B^{i} \ast \sigma\circ\left({f -  A \ast u^{i-1}}\right),$$
where $A$ and $B^i$ are both convolution operators.
Generally, convolutional operators are also used in time series processing, from the frequency perspective, for signal decomposition, especially in classical methods such as wavelet-based methods~(\cite{percival2000wavelet,joo2015time}). 
However, in long-term time series forecasting problems, such as the data sets that we will use in \S\ref{sec:data}, the time series can be highly complex. For example, Figure~\ref{fig:dataset-freq} shows three typical examples in three different data sets, 
which can vary significantly in frequencies and amplitudes. Thus, convolutional operators cannot process
these different data sets uniformly.
A more appropriate choice for the underlying system $A$ is to consider it as a general linear mapping without any special structures.
Consequently, we use a fully connected layer to learn the system $A$. Accordingly, we also choose $B^i$ as a fully connected layer. Thus, we have
\begin{equation}
	u^{\ell,i} = u^{\ell,i-1} + \sigma \circ B^{\ell,i} \circ \left({f^\ell -  A^{\ell} u^{\ell,i-1}}\right), \quad B^{\ell,i}\in\mathbb{R}^{I_\ell \times I_\ell}, A^{\ell}\in\mathbb{R}^{I_\ell \times I_\ell}.
\end{equation}
For simplicity, we use $A^{\ell}\in\mathbb{R}^{I_\ell \times I_\ell}$ and $B^{\ell,i}\in\mathbb{R}^{I_\ell \times I_\ell}$ to denote the fully connected operators with bias, respectively.
Numerical results in Table~\ref{tab:mlpconv} demonstrate that the best combination of $A$ and $B^i$ is the fully connected operator.

\subsection{Multigrid structure: a hierarchical feature extraction scheme}
A hierarchical feature extraction scheme for time series processing is typically standard in classical and CNN-based methods for imaging processing.
For time series signals, the multilevel structure (multi-resolution) can help to capture different features of the signal with different frequencies. For example, N-HiTS~(\cite{challu2022n}) introduced the hierarchical structure into the forecast module in N-BEATS~(\cite{oreshkin2019n}), achieving a higher accuracy with fewer parameters. 

Here, we consider the multigrid structure, which is a more mature and relevant multi-level method to solve linear systems arising from numerical PDEs~(\cite{xu1992iterative,xu2002method,xu2017algebraic}).
In the multigrid method, we first have a series of grids (resolutions) for the feature and data space.
More precisely, let us denote $I_\ell = \frac{I}{2^{\ell-1}}$ as the grid size (resolution) for the $\ell$-th level grid. 
Correspondingly, we have constrained linear models on different grids,
\begin{equation}
	\label{eq:Auf-l}
	A^\ell u^\ell = f^\ell,
\end{equation}
where $u^\ell$, $f^\ell$, and $A^\ell$ represent the feature, data, and underlying system in the $\ell$-th level respectively. By applying the numerical scheme in~\eqref{iterative}, we have the basic block in the $\ell$-th level
\begin{equation}
	u^{\ell,i} = u^{\ell,i-1} + \sigma \circ B^{\ell,i} \sigma  (f^\ell -  A^\ell u^{\ell,i-1}), \quad i=1:\nu_\ell.
\end{equation}
Here, $\nu_\ell$ is the number of iterations on each level, which is a hyper-parameter of the model. Notably, there is only one $A^\ell$ on the $\ell$-th level, while we have $\nu_\ell$ different $B^{\ell,i}$ to learn. This is a fundamental assumption 
of the constrained linear model; specifically, that there can only be one system operator $A^\ell$ on each level, while there can be different solving operators $B^{\ell,i}$ for solving the model $A^\ell u^\ell =f^\ell$.

To integrate different grids, restriction and prolongation operators must be involved, such as different pooling operations in typical CNN models.
Recently, MgNet~(\cite{he2019mgnet,he2021interpretive,wang2022cnns}) has successfully applied the multigrid structure to image classification problems, using convolution with $stride=2$ as the restriction operator to achieve a multi-scale transformation.  The scale transformation for feature $u$ and data $f$ is defined as follows:
\begin{equation}
	\begin{split}
		u^{\ell+1,0} &= \Pi_\ell^{\ell+1}\ast_2u^{\ell,\nu_\ell}\\
		f^{\ell+1} &= R^{\ell+1}_\ell \ast_2 (f^\ell - A^\ell \ast u^{\ell,\nu_\ell}) + A^{\ell+1} \ast u^{\ell+1,0},
	\end{split}
	\label{mg:restri}
\end{equation}
where $\Pi_\ell^{\ell+1}\ast_2$ and $R^{\ell+1}_\ell \ast_2$ are convolution operators with $stride=2$. For the prolongation process in image classification, MgNet uses the deconvolution operator $\Pi_{\ell+1}^{\ell}\ast^2$ with $stride=2$, which interpolates information on a coarse grid to a fine grid.

In our model, considering the fully connected structure, we take the restriction (sub-sampling) and prolongation (interpolation) as linear mappings between $\mathbb R^{I_\ell}$ and $\mathbb R^{I_{\ell+1}}$. More precisely, we take $\Pi_{\ell}^{\ell+1}, R_{\ell}^{\ell+1} \in \mathbb R^{I_{\ell+1} \times I_{\ell}}$ and $\Pi_{\ell+1}^{\ell} \in \mathbb R^{I_{\ell} \times I_{\ell+1}}$ as fully connected layers (with bias) for restoration and prolongation operators.

Another practical reason to introduce the multigrid structure is to further reduce the computational complexity of the proposed model.
Numerical results in Table~\ref{tab:ablation-V-cycle} show the efficiency and improvements obtained by introducing the multigrid structure.

Here, $u=u^{1,\nu_1}$ is the feature extracted by the fully-connected constrained linear model with the multigrid scheme. In addition, we take $W_1\in\mathbb{R}^{I\times O}$ and $W_2\in\mathbb{R}^{O\times O}$ as the fully connected layers (with bias) to construct the one-hidden layer approximation for the feature interpolation $\mathcal I(u)$. 

\subsection{FV-MgNet: A fully connected V-cycle MgNet model for time series forecasting}
By combining the feature extraction process discussed above and the feature interpolation approximation in~\eqref{eq:fI}, we have the FV-MgNet algorithm for time series forecasting.
\begin{algorithm}[H]
	\caption{$y=\text{FV-MgNet}(f, J, \nu_\ell, I, O)$}
	\label{alg:mgnet}
	\begin{algorithmic}[1]
		\STATE {\bf Input}: series $f\in \mathbb{R}^{I}$, number of grids J, number of smoothing iterations $\nu_\ell$ for $\ell=1:J$,
		input length $I$, output length $O$.
		\STATE Initialization:  $f^1 = Wf$,~$W\in\mathbb{R}^{I\times I}$, ~$u^{1,0}=0$,~$I_\ell = \frac{I}{2^{\ell-1}}$
		\FOR{$\ell = 1:J$}
		\FOR{$i = 1:\nu_\ell$}
		\STATE Feature extraction (smoothing):
		\begin{equation*}\label{eq:mgnet}
			u^{\ell,i} = u^{\ell,i-1} + \sigma \circ B^{\ell,i} \circ \left({f^\ell -  A^{\ell} u^{\ell,i-1}}\right), \quad B^{\ell,i}\in\mathbb{R}^{I_\ell \times I_\ell},A^{\ell}\in\mathbb{R}^{I_\ell \times I_\ell}
		\end{equation*}
		\ENDFOR
		\STATE Note:
		$
		u^\ell= u^{\ell,\nu_\ell}
		$
		\IF{$\ell<J$}
		\STATE Restriction:
		\begin{equation*}
			\label{eq:interpolation}
			u^{\ell+1,0} = \Pi_\ell^{\ell+1}u^{\ell},\quad \Pi_\ell^{\ell+1}\in\mathbb{R}^{I_{\ell+1}\times I_\ell}
		\end{equation*}
		\begin{equation*}
			\label{eq:restrict-f}
			f^{\ell+1} = R^{\ell+1}_\ell (f^\ell - A^\ell  u^{\ell}) + A^{\ell+1}  u^{\ell+1,0},\quad  R_\ell^{\ell+1}\in\mathbb{R}^{I_{\ell+1}\times I_\ell}
		\end{equation*}
		\ENDIF
		\ENDFOR
		\FOR{$\ell = J-1:1$}
		\STATE Prolongation
		\begin{equation*}
			u^{\ell,0}=u^{\ell}+\Pi_{\ell+1}^{\ell}(u^{\ell+1}-u^{\ell+1,0}), \quad \Pi_{\ell+1}^{\ell}\in\mathbb{R}^{I_{\ell} \times I_{\ell+1}}
		\end{equation*}
		\FOR{$i = 1:\nu_\ell$}
		\STATE Feature extraction (smoothing):
		\begin{equation*}
			u^{\ell,i} = u^{\ell,i-1} + \sigma \circ \overline{B}^{\ell,i}\circ\left({f^\ell -  \overline{A}^{\ell} u^{\ell,i-1}}\right), \quad \overline{B}^{\ell,i}\in\mathbb{R}^{I_\ell \times I_\ell},\overline{A}^{\ell}\in\mathbb{R}^{I_\ell \times I_\ell}
		\end{equation*}
		\ENDFOR
		\ENDFOR\\
		{Feature Interpolation:}
		$$y=W^2\sigma\left( W^1u^{1,\nu_1}\right), \quad u^{1,\nu_1}\in\mathbb{R}^{I}, ~ W_1\in\mathbb{R}^{I\times O}, ~ W_2\in\mathbb{R}^{O\times O}$$
	\end{algorithmic}
\end{algorithm}

\section{Experiments}\label{experiments}
To verify the performance of the FV-MgNet, we conducted experiments on six popular real-world data sets, spanning  the domains electricity, economy, traffic, weather, and disease. The models we compared included Transformer models, such as FEDformer~(\cite{zhou2022fedformer}), ETSformer~(\cite{woo2022etsformer}), Autofomer~(\cite{Autoformer}),  the fully connected layer-based model N-HiTS~(\cite{challu2022n}), and the frequency enhanced model FiLM~(\cite{zhou2022film}).

\subsection{Datasets}\label{sec:data}

We considered six benchmark data sets across diverse domains. The details of these data sets are summarized as follows. 1) The ETT\footnote{https://github.com/zhouhaoyi/ETDataset} (\cite{haoyietal-informer-2021}) data set contains two sub-datasets (ETT1 and ETT2) collected from two electricity transformers; there are two versions of each data set at different resolutions (15min and 1h), with multiple series of loads and one series of oil temperatures. 2) The Electricity\footnote{https://archive.ics.uci.edu/ml/datasets/ElectricityLoadDiagrams20112014} data set contains the electricity consumption of clients, with each column corresponding to one client. 3) The Exchange\footnote{https://github.com/laiguokun/multivariate-time-series-data} (\cite{lai2018modeling}) data set contains the exchange of eight countries. 4) The Traffic\footnote{http://pems.dot.ca.gov} dataset contains the occupation rate of the freeway system across California (United States). 5) The Weather\footnote{https://www.bgc-jena.mpg.de/wetter/} dataset contains 21 meteorological indicators for a range of one year in Germany. 6) the Illness\footnote{https://gis.cdc.gov/grasp/fluview/fluportaldashboard.html} dataset contains the influenza-like illness patients in the United States, the detail of data sets show in Table \ref{tab:dataset}.

\begin{table}[htbp]
	\caption{Summarized feature details of eight datasets.}
	\label{tab:dataset}
	\begin{center}
		\begin{sc}
			\scalebox{0.95}{
				\begin{tabular}{l|cccr}
					\toprule
					Dataset & len & dim & freq \\
					\midrule
					ETTm2 & 69680 & 7 & 15 min\\
					Electricity & 26304 & 321 & 1h & \\
					Exchange & 7588 & 8 & 1 day\\
					Traffic & 17544 & 862 & 1h & \\
					Weather & 52696 & 21 & 10 min & \\
					ILI & 966 & 8 & 7 days\\
					\bottomrule
				\end{tabular}
			}
		\end{sc}
\end{center}
\vskip -0.1in
\end{table}

\subsection{Experiment settings}
We used mean squared error (MSE) and mean absolute error (MAE) to evaluate the performance of our model, the most widely used in previous works, for the predict length $H$,
\begin{equation}
MSE = \frac{1}{H}\sum^{t+H}_{\tau=t}(y_{\tau}-\hat{y}_{\tau})^2\quad\quad\quad MAE = \frac{1}{H}\sum^{t+H}_{\tau=t}\left|y_{\tau}-\hat{y}_{\tau}\right|
\end{equation}

The data sets were split into train, validation, and test sets chronologically, following a 70/10/20 split for all the data sets except for ETTm2, which followed a 60/20/20 split (as per convention). Data pre-processing was performed by standardization based on train set statistics.

In the experimental setting, the output lengths were set to 24, 36, 48, and 60 for the ILI data set and 96, 192, 336, and 720 for the other data sets. The input length in all the Transformer models was fixed to 36 for the ILI data set and 96 for the other data sets. N-HiTS and FiLM determine the input length by tuning the parameters and investigating their performance in different data sets. To fairly compare the performance of these models, we used two ways to select the input length: 1) the input length was fixed at 96, denoted as FV-MgNet (96), which has the same settings as the Transformer; and 2) the input length was fixed at one of 96, 192, 336, or 720 for each data set, denoted as FV-MgNet. This latter  method can better demonstrate the advantages of FV-MgNet.

\subsection{Experiment results}
The overall performance of FV-MgNet is summarized in Figure~\ref{fig:mse-param} comparing with different models on the multivariate Electricity data set. The mean squared errors of FV-MgNet perform much better than other benchmarks with different output lengths, especially when the output length is large. Furthermore, the total number of parameters of FV-MgNet is only $15\%$ of those of the Transformer-type models and N-HiTS, and $25\%$ of FiLM as shown in the right plot of Figure~\ref{fig:mse-param}. In the following, we investigate the the performance of FV-MgNet from different perspectives.

\textbf{Forecasting result} As shown in Table \ref{tab:multivariate}, FV-MgNet achieves the best performance on the six data sets when compared with the other models, except for FiLM on Exchange and Traffic. Compared with ETSformer, which achieves state-of-the-art results in Transformer type models, FV-MgNet reduces MSE error by more than $18\%$; improved results were particularly obvious on the traffic and ILI data sets, in which MSE decreases of $31\%$ and $29\%$ MSE, respectively, were observed. Compared with N-HiTS, which uses the fully connected layer as the basic block,  FV-MgNet reduces MSE by $8\%$. Although FV-MgNet and  FiLM have a comparable performance on Exchange and Traffic, FV-MgNet outperforms FiLM by $5\%$ MSE and $6\%$ MAE on the other data sets. Notably, FV-MgNet achieves the best results in the longest time-series prediction of all datasets (ILI with an output length 60 and the other data sets with an output length of 720), which indicates our model is more suitable for long-range prediction compared to other models. 
\begin{table}[H]
\centering
\caption{Multivariate results with different prediction lengths $O \in\{96,192,336,720\} .$ A lower MSE or MAE indicates a better prediction. The best results are highlighted in bold.}\vspace{-3mm}
\vskip 0.1in
\begin{scriptsize}
	\begin{tabular}{c|c|cccccccccccccccccc}
		\toprule
		\multicolumn{2}{c|}{Methods}&\multicolumn{2}{c|}{FV-MgNet}&\multicolumn{2}{c|}{FiLM}&\multicolumn{2}{c|}{N-HiTS}&\multicolumn{2}{c|}{ETSformer}&\multicolumn{2}{c|}{FEDformer}&\multicolumn{2}{c}{Autoformer}\\
		\midrule
		\multicolumn{2}{c|}{Metric}&MSE & MAE& MSE & MAE& MSE & MAE& MSE & MAE& MSE & MAE&MSE  & MAE\\
		\midrule
		\multirow{4}{*}{\rotatebox{90}{$ETTm2$}}
		& 96  & 0.173 &\textbf{0.253}& \textbf{0.165}& 0.256 & 0.176& 0.255& 0.189 & 0.280& 0.203& 0.287 & 0.255& 0.339\\
		& 192 &0.230 &\textbf{0.296} & \textbf{0.222}& \textbf{0.296} & 0.245& 0.305& 0.253 & 0.319& 0.269& 0.328 & 0.281& 0.340\\
		& 336 &0.279 &\textbf{0.329}& \textbf{0.277}&0.333 & 0.295& 0.346& 0.314 & 0.357& 0.325& 0.366 & 0.339& 0.372\\
		& 720 &\textbf{0.367} &\textbf{0.385}& 0.371& 0.389 & 0.401& 0.426& 0.414 &0.413& 0.421& 0.415 & 0.422& 0.419\\
		\midrule
		\multirow{4}{*}{\rotatebox{90}{$Electricity$}}
		& 96  &\textbf{0.144} &\textbf{0.250} & 0.154& 0.267 & 0.147& 0.249    & 0.187& 0.304 & 0.183& 0.297 & 0.201& 0.317\\
		& 192 &\textbf{0.163} &0.262& 0.164& \textbf{0.258} & 0.167& 0.269   & 0.199& 0.315 & 0.195& 0.308 & 0.222& 0.334\\
		& 336 &\textbf{0.176} &\textbf{0.276}& 0.188& 0.283 & 0.186& 0.290   & 0.212& 0.329& 0.212& 0.313 & 0.231& 0.338\\
		& 720 &\textbf{0.212} &\textbf{0.308}& 0.236& 0.338 & 0.243& 0.340  & 0.233& 0.345& 0.231& 0.343 & 0.254& 0.361\\
		\midrule
		\multirow{4}{*}{\rotatebox{90}{$Exchange$}}
		& 96  &0.082 &0.206 & \textbf{0.079}&\textbf{0.204} & 0.092& 0.211    & 0.085& 0.204& 0.139& 0.276 & 0.197& 0.323\\
		& 192 &0.184 &0.314& \textbf{0.159}& \textbf{0.292} & 0.208& 0.322    & 0.182& 0.303& 0.256& 0.369 & 0.300& 0.369\\
		& 336 &0.307 &0.416& \textbf{0.270}& \textbf{0.398} & 0.371& 0.443  & 0.348& 0.428& 0.426& 0.464 & 0.509& 0.524\\
		& 720 &\textbf{0.554} &\textbf{0.582} & 0.830& 0.721& 0.888& 0.723   & 1.025& 0.774& 1.090& 0.800 & 1.447& 0.941\\
		\midrule
		\multirow{4}{*}{\rotatebox{90}{$Traffic$}} 
		&96    &\textbf{0.396} &\textbf{0.285}  &0.416  &0.294 &0.402  &0.282    &0.607  &0.392   &0.562  &0.349 &0.613  &0.388  \\
		& 192  &0.417 &0.295   &\textbf{0.408}  &\textbf{0.288} &0.420  &0.297    &0.621  &0.399 &0.562  &0.346 &0.616  &0.382   \\
		& 336  & 0.436 &0.302 &\textbf{0.425}  &\textbf{0.298} &0.448  &0.313   &0.622&0.396&0.570  &0.323 &0.622  &0.337  \\
		& 720  &\textbf{0.468} &\textbf{0.315} &0.520  &0.353 &0.539  &0.353    &0.632  &0.396&0.596  &0.368 &0.660  &0.408 \\
		\midrule
		\multirow{4}{*}{\rotatebox{90}{$Weather$}} 
		& 96  & \textbf{0.155} & \textbf{0.196} &0.199  &0.262 &0.158  &0.195  &0.197 &0.281  &0.217  &0.296 &0.266  &0.336   \\
		& 192 & \textbf{0.201} & \textbf{0.239} &0.228  &0.288 &0.211  &0.247  &0.237 &0.312  &0.276  &0.336 &0.307 &0.367  \\
		& 336 & \textbf{0.244} & \textbf{0.279} &0.267  &0.323 &0.274  &0.300  &0.298 &0.353  &0.339  &0.380 &0.359  &0.395   \\
		& 720 & \textbf{0.313} & \textbf{0.329} &0.319  &0.361 &0.351  &0.353  &0.352 &0.388  &0.403  &0.428 &0.419  &0.428  \\
		\midrule
		\multirow{4}{*}{\rotatebox{90}{$ILI$}} 
		&24 &\textbf{1.647} &\textbf{0.764} &1.970  &0.875 &1.862  &0.869 &2.527 &1.020 &2.203  &0.963 &3.483 &1.287 \\
		&36 &\textbf{1.841} &\textbf{0.839} &1.982  &0.859 &2.071  &0.969 &2.615 &1.007 &2.272  &0.976 &3.103 &1.148 \\
		&48 &\textbf{1.831} &\textbf{0.853} &1.868  &0.896 &2.346  &1.042 &2.359 &0.972 &2.209  &0.981 &2.669 &1.085  \\
		&60 &\textbf{1.765} &\textbf{0.814} &2.057  &0.929 &2.560  &1.073 &2.487 &1.016 &2.545  &1.061 &2.770 &1.125 \\
		\bottomrule
	\end{tabular}
\end{scriptsize}
\label{tab:multivariate}
\vskip -0.1in
\end{table}

\begin{figure}[htbp]
\centering
\begin{subfigure}[b]{\textwidth}
	\centering
	\includegraphics[width=\textwidth]{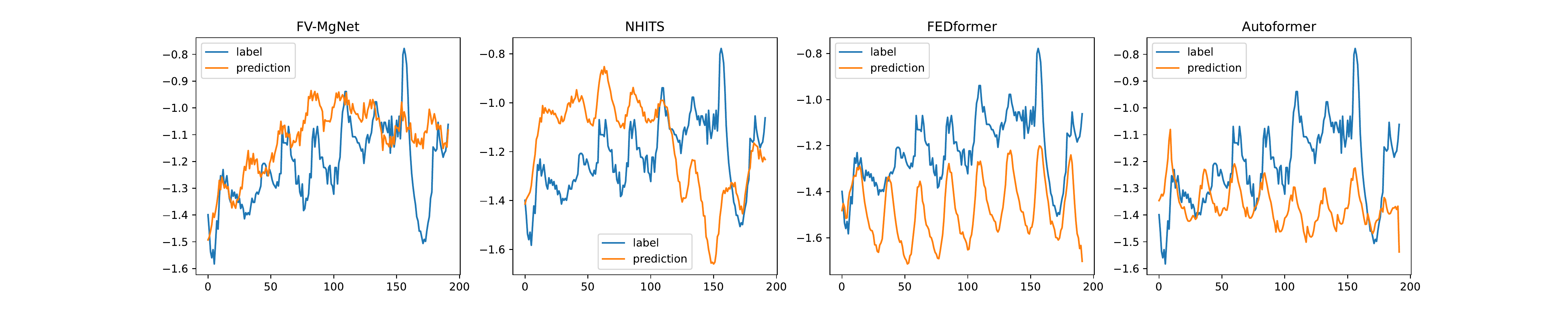}
	\caption{$ETTh1$}
	
\end{subfigure}
\quad
\begin{subfigure}[b]{\textwidth}
	\centering
	\includegraphics[width=\textwidth]{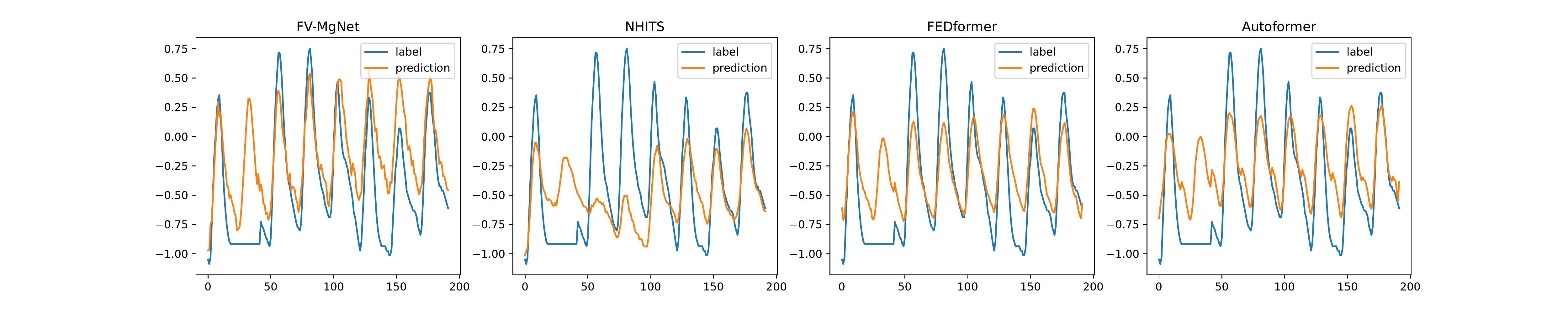}
	\caption{$ETTh2$}
	
\end{subfigure}
\quad
\begin{subfigure}[b]{\textwidth}
	\centering
	\includegraphics[width=\textwidth]{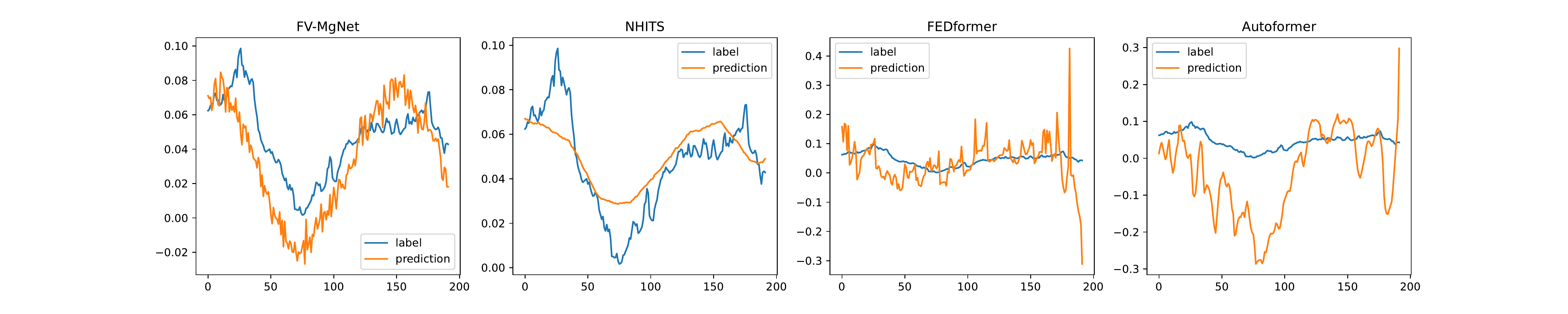}
	\caption{$Weather$}
	\label{fig:five over x}
\end{subfigure}
\caption{ Visualization of the forecasting output (Y-axis) of four models (FV-MgNet, N-HiTS, FEDformer, and Autoformer) with an output length $O=192$ (X-axis) on the ETTh1, ETTh2,and Weather data sets, respectively.}
\label{fig:results-show}
\end{figure}

\textbf{Prediction results visualization} In Figure \ref{fig:results-show}, we plot the prediction results of four models (FV-MgNet, N-HiTS, FEDformer, and Autoformer) on three data sets with different temporal patterns: ETTh1 (testset sequence 650, variable 7), ETTh2 (testset sequence 150, variable 7), and Weather (testset sequence 185, variable 20). On ETTh1, FV-MgNet can accurately predict seasonal and trend information and N-HiTS can partially capture trend information, whereas FedFormer and Autoformer can only capture seasonal information. In ETTh2, all four methods can accurately capture season information, and FV-MgNet can capture trend information more accurately than the others. In Weather, FV-MgNet can predict both season and trend information accurately, N-HiTS can partially capture trend information, whereas FEDformer and Autofomer cannot predict either season or  trend information.

\textbf{Computational efficiency}  We recorded the memory usage and running time of each epoch during the training process in Figure \ref{fig:memory-time}, where we set the input length to 96 and changed different output lengths. Both memory usage and running time of FV-MgNet are the smallest. The running time of FV-MgNet is ten times faster than Autoformer and two times faster than FiLM. The running time of FV-MgNet is ten times faster than Autoformer, two times faster than FiLM, and $10\%$ faster than N-HiTS. The memory usage of FV-MgNet is five times less than Autoformer, two times less than FiLM, and 1.5 times less than N-HiTS. Furthermore, the memory usage and running time of FV-MgNet are almost quasi-constant as the output length increases.

\textbf{Robustness} Intuitively, if the output length is fixed, better results should be otained from a longer input length. We set the input length to 96 and increased it from 96 to 720.  As shown in Table \ref{tab:prolong-input-length}, the prediction results are reduced when we increase the input length of the Transformer type models. However, FV-MgNet maintains stable prediction results on the ETT data set and even obtains a significant improvement on the Weather data set. The results show that FV-MgNet can effectively capture the features in long input sequences, and the robustness of FV-MgNet is significantly better than Transformer models in long input sequences.
\begin{table}[htbp]
\centering
\caption{ The MSE comparisons in robustness experiment of forecasting $96$ steps for the $ETTm2$ and weather data sets with prolonged input length.}\vspace{-3mm}
\vskip 0.1in
\setlength\tabcolsep{5pt}
\small
	\begin{tabular}{c|c|cccccccccc}
		\toprule
		\multicolumn{2}{c|}{Methods}&FV-MgNet & FEDformer & Autoformer & Informer & Reformer & LogFormer & Transformer\\
		\midrule
		\multicolumn{2}{c|}{Metric} & MSE & MSE & MSE& MSE& MSE& MSE& MSE\\
		\hline
		\multirow{4}{*}{\rotatebox{90}{$ETTm2$}}
		& 96  &0.176 & 0.203 & 0.239 & 0.428& 0.615 & 0.667& 0.557\\
		& 192 &0.173 & 0.275 & 0.265 & 0.385& 0.686 & 0.697& 0.710\\
		& 336 &0.173 & 0.347 & 0.375 & 1.078& 1.359 & 0.937& 1.078\\
		& 720 &0.180 & 0.392 & 0.315 & 1.057& 1.443 & 2.153& 1.691\\
		\midrule
		\multirow{4}{*}{\rotatebox{90}{$Weather$}}
		& 96  &0.185 & 0.217 & 0.266 & 0.300 & 0.689 & 0.458& 0.604\\
		& 192 &0.167 & 0.253 & 0.276 & 0.325 & 0.724 & 0.537& 0.674\\
		& 336 &0.155 & 0.278 & 0.336 & 0.346 & 0.736 & 0.742& 0.692\\
		& 720 &0.153 & 0.276 & 0.547 & 0.421 & 0.701 & 0.884& 0.740\\
		\bottomrule
	\end{tabular}				
\label{tab:prolong-input-length}
\vskip -0.1in
\end{table}

\subsection{Ablation results}
In this subsection, we investigated the structure of MgNet for forecasting problems. We first tested two different operators, convolutional operators or fully connected operators, for A and B under V-cycle structure MgNet. The results in Table \ref{tab:mlpconv} show that both A and B, using the fully connected operators, are significantly better than the other choices. This verifies that fully connected layers are more suitable for time series forecasting tasks than convolution layers under the MgNet framework. Next, we explored the structure of fully connected MgNet. We designed three models: (1) Residual network, which uses the same dimension for A and B in different grids (the algorithm is shown in Algorithm \ref{alg:residual} ); (2) {$\backslash$-MgNet}, which has only a single hierarchical structure (the algorithm is shown in Algorithm \ref{alg:backslashmgnet} ); and 
(3) FV-MgNet as shown in Algorithm \ref{alg:mgnet}. (3) FV-MgNet as shown in Algorithm \ref{alg:mgnet}. The results obtained using the residual network are poor but can be improved as a $\backslash$-MgNet after adding a one-side hierarchical structure. However, numerical results demonstrate that FV-MgNet, MgNet with V-cycle (two-side hierarchical structure), is the best choice. 
\begin{table}[htbp]
\centering
\caption{Ablation experiments to compare FV-MgNet, $\backslash$-MgNet, and Residual model. Our numerical results demonstrate that multigrid structure for feature extraction can indeed improve the accuracy of the method.}\vspace{-3mm}
\vskip 0.1in
	\small
	\begin{tabular}{c|c|cccccccc}
		\toprule
		\multicolumn{2}{c|}{Methods}&\multicolumn{2}{c|}{FV-MgNet}&\multicolumn{2}{c|}{$\backslash$-MgNet}&\multicolumn{2}{c}{Residual}\\
		\midrule
		\multicolumn{2}{c|}{Metric} & MSE & MAE& MSE & MAE&MSE  & MAE\\
		\midrule
		\multirow{4}{*}{\rotatebox{90}{$ETTm2$}}
		& 96  &\textbf{0.173} &\textbf{0.253} & 0.183& 0.270 & 0.212& 0.318\\
		& 192 &\textbf{0.230} &\textbf{0.296} & 0.247& 0.311 & 0.275& 0.340\\
		& 336  &\textbf{0.279} &\textbf{0.329}& 0.321& 0.356 & 0.354& 0.385\\
		& 720 &\textbf{0.367} &\textbf{0.385} & 0.434& 0.422 & 0.482& 0.462\\
		\midrule
		\multirow{4}{*}{\rotatebox{90}{$Electricity$}}
		& 96  &\textbf{0.144} &\textbf{0.250} & 0.179& 0.287 & 0.213& 0.312\\
		& 192 &\textbf{0.163} &\textbf{0.262} & 0.204& 0.305 & 0.262& 0.349\\
		& 336 &\textbf{0.176} &\textbf{0.276} & 0.216& 0.317 & 0.272& 0.370\\
		& 720 &\textbf{0.212} &\textbf{0.308} & 0.280& 0.369 & 0.341& 0.436\\
		\midrule
		\multirow{4}{*}{\rotatebox{90}{$Exchange$}}
		& 96  &\textbf{0.082} &\textbf{0.206} & 0.093& 0.223 & 0.104& 0.283\\
		& 192 &\textbf{0.184} &\textbf{0.314} & 0.242& 0.361 & 0.298& 0.394\\
		& 336 &\textbf{0.307} &\textbf{0.416} & 0.358& 0.455 & 0.443& 0.552\\
		& 720 &\textbf{0.554} &\textbf{0.582} & 1.258& 0.834 & 1.430& 0.968\\
		\midrule
		\multirow{4}{*}{\rotatebox{90}{$Traffic$}} 
		&96 &\textbf{0.396} &\textbf{0.285}     &0.717  &0.398 &0.589  &0.348\\
		& 192 &\textbf{0.417}  &\textbf{0.295}  &0.686  &0.401 &0.567  &0.345   \\
		& 336  & \textbf{0.436} &\textbf{0.302} &0.661  &0.388 &0.553  &0.342  \\
		& 720 &\textbf{0.468} &\textbf{0.315}   &0.666  &0.384 &0.612  &0.351  \\
		\midrule
		\multirow{4}{*}{\rotatebox{90}{$Weather$}} 
		&96 &\textbf{0.155} &\textbf{0.196}     &0.181  &0.222 &0.217  &0.263   \\
		& 192 &\textbf{0.201}  &\textbf{0.239}  &0.218  &0.259 &0.255 &0.308 \\
		& 336  & \textbf{0.244} &\textbf{0.279} &0.267 &0.303 &0.302  &0.331   \\
		& 720 &\textbf{0.313} &\textbf{0.329}   &0.335  &0.345 &0.389  &0.403   \\
		\midrule
		\multirow{4}{*}{\rotatebox{90}{$ILI$}} 
		&24 &\textbf{1.647} &\textbf{0.764}   &3.123  &1.130 &3.261  &1.280  \\
		&36 &\textbf{1.841}  &\textbf{0.839}  &3.247  &1.182 &3.538 &1.461  \\
		&48  & \textbf{1.831} &\textbf{0.853} &2.890  &1.063 &3.295  &1.198   \\
		&60 &\textbf{1.765} &\textbf{0.814}   &3.218  &1.153 &3.418  &1.229  \\
		\bottomrule
	\end{tabular}
\label{tab:ablation-V-cycle}
\end{table}

\begin{algorithm}[htbp]
\caption{$y=\text{Residual}(f,\nu, I, O)$}
\label{alg:residual}
\begin{algorithmic}[1]
	\STATE {\bf Input}: series $f\in \mathbb{R}^{I}$, number of smoothing iterations $\nu$,
	input length $I$, output length $O$.
	\STATE Initialization:  $f^1 = Wf$,~$W\in\mathbb{R}^{I\times I}$, ~$u^{0}=0$
	\FOR{$i = 1:\nu$}
	\STATE Feature extraction (smoothing):
	\begin{equation*}
		u^{i} = u^{i-1} + \sigma \circ B^{i}\left({f -  A u^{i-1}}\right), \quad B^{i}\in\mathbb{R}^{I \times I},A\in\mathbb{R}^{I\times I}
	\end{equation*}
	\ENDFOR\\
	{Feature Interpolation:}
	$$y=W^2\sigma\left( W^1(u^{\nu}) +b^1\right), \quad u^{\nu}\in\mathbb{R}^{I}, ~ W_1\in\mathbb{R}^{I\times O}, ~ W_2\in\mathbb{R}^{O\times O}$$
\end{algorithmic}
\end{algorithm}

\begin{algorithm}[htbp]
\caption{$y=\text{$\backslash-$MgNet}(f, J, \nu_\ell, I, O)$}
\label{alg:backslashmgnet}
\begin{algorithmic}[1]
	\STATE {\bf Input}: series $f\in \mathbb{R}^{I}$, number of grids J, number of smoothing iterations $\nu_\ell$ for $\ell=1:J$,
	input length $I$, output length $O$.
	\STATE Initialization:  $f^1 = Wf$,~$W\in\mathbb{R}^{I\times I}$, ~$u^{1,0}=0$,~$I_\ell = \frac{I}{2^{\ell-1}}$
	\FOR{$\ell = 1:J$}
	\FOR{$i = 1:\nu_\ell$}
	\STATE Feature extraction (smoothing):
	\begin{equation*}
		u^{\ell,i} = u^{\ell,i-1} + \sigma \circ B^{\ell,i}\left({f^\ell -  A^{\ell} u^{\ell,i-1}}\right), \quad B^{\ell,i}\in\mathbb{R}^{I_\ell \times I_\ell},A^{\ell}\in\mathbb{R}^{I_\ell \times I_\ell}
	\end{equation*}
	\ENDFOR
	\STATE Note:
	$
	u^\ell= u^{\ell,\nu_\ell}
	$
	\IF{$\ell<J$}
	\STATE Restriction:
	\begin{equation*}
		u^{\ell+1,0} = \Pi_\ell^{\ell+1}u^{\ell},\quad \Pi_\ell^{\ell+1}\in\mathbb{R}^{I_{\ell+1}\times I_\ell}
	\end{equation*}
	\begin{equation*}
		f^{\ell+1} = R^{\ell+1}_\ell (f^\ell - A^\ell  u^{\ell}) + A^{\ell+1}  u^{\ell+1,0},\quad  R_\ell^{\ell+1}\in\mathbb{R}^{I_{\ell+1}\times I_\ell}
	\end{equation*}
	\ENDIF
	\ENDFOR\\
	{Feature Interpolation:}
	$$y=W^2\sigma\left(W^1 u^{J,\nu_J}\right), \quad u^{1,\nu_1}\in\mathbb{R}^{I_J}, ~ W_1\in\mathbb{R}^{I_J\times O}, ~ W_2\in\mathbb{R}^{0\times O}$$
\end{algorithmic}
\end{algorithm}

\begin{table}[htbp]
\centering
\caption{Different encoding methods for A and B, marked as (A,B), the first element represents the A encoding method and the second element represents the B encoding method. The first row represents the selected encoding method, where Conv represents the convolution layer, and FC represents the fully connected layer.  }\vspace{-3mm}
\vskip 0.1in
	\small
	\begin{tabular}{c|c|cccccccccc}
		\toprule
		\multicolumn{2}{c|}{(A,B)}&\multicolumn{2}{c|}{(FC,FC)}&\multicolumn{2}{c|}{(Conv,Conv)}&\multicolumn{2}{c|}{(Conv,FC)}&\multicolumn{2}{c}{(FC,Conv)}\\
		\midrule
		\multicolumn{2}{c|}{Metric} & MSE & MAE& MSE & MAE & MSE & MAE& MSE & MAE\\
		\midrule
		\multirow{4}{*}{\rotatebox{90}{$ETTm2$}}
		& 96  &\textbf{0.173} &\textbf{0.253} & 0.464& 0.383 & 0.304 & 0.379 & 0.355 & 0.351\\
		& 192 &\textbf{0.230} &\textbf{0.296} & 0.608& 0.610 & 0.438 & 0.540 & 0.448 & 0.366\\
		& 336  &\textbf{0.279} &\textbf{0.329}& 0.893& 1.276 & 0.481 & 0.624 & 0.588 & 0.487\\
		& 720 &\textbf{0.367} &\textbf{0.385} & 1.151& 2.350 & 0.824 & 1.319 & 0.737 & 0.628\\
		\midrule
		\multirow{4}{*}{\rotatebox{90}{$Electricity$}}
		& 96  &\textbf{0.144} &\textbf{0.250} & 0.407& 0.336 & 0.340 & 0.418 & 0.349 & 0.423 \\
		& 192 &\textbf{0.163} &\textbf{0.262} & 0.410& 0.336 & 0.335 & 0.420 & 0.447 & 0.498\\
		& 336 &\textbf{0.176} &\textbf{0.276} & 0.393& 0.316 & 0.460 & 0.507 & 0.423 & 0.493\\
		& 720 &\textbf{0.212} &\textbf{0.308} & 0.409& 0.335 & 0.384 & 0.457 & 0.379 & 0.447\\
		\midrule
		\multirow{4}{*}{\rotatebox{90}{$Exchange$}}
		& 96  &\textbf{0.082} &\textbf{0.206} & 0.798& 0.875 & 0.492 & 0.348 & 0.528 & 0.366\\
		& 192 &\textbf{0.184} &\textbf{0.314} & 0.977& 1.364 & 0.615 & 0.447 & 0.428 & 0.474\\
		& 336 &\textbf{0.307} &\textbf{0.416} & 0.903& 1.180 & 0.676 & 0.482 & 0.486 & 0.493\\
		& 720 &\textbf{0.554} &\textbf{0.582} & 1.873& 2.469 & 1.506 & 1.092 & 1.629 & 1.086\\
		\midrule
		\multirow{4}{*}{\rotatebox{90}{$Weather$}} 
		& 96  &\textbf{0.155} &\textbf{0.196} &0.256  &0.174 & 0.174 & 0.258 & 0.168 & 0.251\\
		& 192 &\textbf{0.201} &\textbf{0.239} &0.308  &0.243 & 0.226 & 0.309 & 0.225 & 0.305 \\
		& 336 &\textbf{0.244} &\textbf{0.279} &0.351  &0.289 & 0.321 & 0.362 & 0.296 & 0.365 \\
		& 720 &\textbf{0.313} &\textbf{0.329} &0.401  &0.366 & 0.585 & 0.490 & 0.482 & 0.454\\
		\bottomrule
	\end{tabular}
\label{tab:mlpconv}
\vskip -0.1in
\end{table}

\subsection{Other Studies}
Details of other experimental studies are provided in bellow: (1)  univariate forecasting experiments, (2) Transformer models experiments under the same setting, (3) parameter sensitivity experiments of grid number and iteration number, (4) experiments under different input lengths, and (5) ETT full benchmark experiments.

\subsubsection{Univariate forecasting results}
We tested the two data sets in a univariate setting and the results are shown in Table \ref{tab:univariate}. Compared with the extensive baselines, FV-MgNet achieves competitive results, especially in long-term forecasting on ETTm2. FV-MgNet can reduce MSE error by 12\%.
\begin{table}[htbp]
\centering
\caption{Univariate forecasting results with different prediction lengths $O \in\{96,192,336,720\} .$ A lower MSE or MAE indicates a better prediction. The best results are highlighted in bold.}\vspace{-3mm}
\vskip 0.1in
\setlength\tabcolsep{5pt}
	\small
	\begin{tabular}{c|c|cccccccccccc}
		\toprule
		\multicolumn{2}{c|}{Methods}&\multicolumn{2}{c|}{FV-MgNet}&\multicolumn{2}{c|}{FiLM}&\multicolumn{2}{c|}{N-HiTS}&\multicolumn{2}{c|}{ETSformer}&\multicolumn{2}{c|}{FEDformer}&\multicolumn{2}{c}{Autoformer}\\
		\midrule
		\multicolumn{2}{c|}{Metric} & MSE & MAE & MSE & MAE & MSE & MAE & MSE & MAE & MSE & MAE & MSE & MAE\\
		\midrule
		\multirow{4}{*}{\rotatebox{90}{$ETTm2$}}
		& 96  &0.077 & 0.213 &0.065 & 0.189 & 0.066 & \textbf{0.185} & 0.080 & 0.212 & \textbf{0.063} & 0.189 & 0.065 & 0.189\\
		& 192 &0.101 & 0.247 &0.094 & 0.233 & \textbf{0.087} & \textbf{0.223} & 0.150 & 0.302 & 0.102 & 0.245 & 0.118 & 0.256\\
		& 336 &0.121 & 0.272 &0.121 & 0.274 & \textbf{0.106} & \textbf{0.251} & 0.175 & 0.334 & 0.130 & 0.279 & 0.154 & 0.305\\
		& 720 &\textbf{0.153} & \textbf{0.310} &0.173 & 0.323 & 0.157 & 0.312 & 0.224 & 0.379 & 0.178 & 0.325 & 0.182 & 0.335\\
		\midrule
		\multirow{4}{*}{\rotatebox{90}{$Exchange$}}
		& 96  &0.123 & 0.245 &0.110 & 0.259 & \textbf{0.093} & \textbf{0.223} & 0.099 & 0.230 & 0.131 & 0.284 & 0.241 & 0.387\\
		& 192 &0.295 & 0.402 &\textbf{0.207} & 0.352 & 0.230 & \textbf{0.313} & 0.223 & 0.353 & 0.277 & 0.420 & 0.300 & 0.369\\
		& 336 &0.337 & 0.486 &\textbf{0.327} & \textbf{0.461} & 0.370 & 0.486 & 0.421 & 0.497 & 0.426 & 0.511 & 0.509 & 0.524\\
		& 720 &\textbf{0.641} & 0.663 &0.811 & 0.708 & 0.728 & \textbf{0.569} & 1.114 & 0.807 & 1.162 & 0.832 & 1.260 & 0.867\\
		\bottomrule
	\end{tabular}				
\label{tab:univariate}
\vskip -0.2in
\end{table}

\subsubsection{Comparison with Transformer models under the same settings}
Transformer-type models set the input length I as 36 for ILI and 96 for the others. We also experimented with FV-MgNet under the same settings, and  we called this model FV-MgNet(96); the baseline of Transformer models contain FEDformer~(\cite{zhou2022fedformer}), ETSformer~(\cite{woo2022etsformer}), Autofomer~(\cite{Autoformer}), Informer~(~\cite{haoyietal-informer-2021}), LogTransfomer~(\cite{Log-transformer-shiyang-2019}), Reformer~(\cite{DBLP:conf/iclr/KitaevKL20-reformer}), the results are shown in Table \ref{tab:96multivariate}.  Compared with ETSformer, FV-MgNet decreases MSE error by more than $8\%$ and MAE error by $11\%$. On the Exchange, Traffic and ILI data sets, the improvement was particularly obvious, with decreases in MSE error of $15\%$, $10\%$, and $25\%$ respectively. The experimental results show that FV-MgNet can also well capture the information in time series under the short input length.

\begin{table}[htbp]
\centering
\caption{Multivariate results with different prediction lengths $O \in\{96,192,336,720\} .$ A lower MSE or MAE indicates a better prediction. We set the
	input length $I$ as 36 for ILI and 96 for the others.  The best results are highlighted in bold.}\vspace{-3mm}
\vskip 0.1in
\begin{scriptsize}
		\begin{tabular}{c|c|cccccccccccccccccccccccc}
			\toprule
			\multicolumn{2}{c|}{Methods}&\multicolumn{2}{c|}{FV-MgNet(96)}&\multicolumn{2}{c|}{ETSformer}&\multicolumn{2}{c|}{FEDformer}&\multicolumn{2}{c|}{Autoformer}&\multicolumn{2}{c|}{Informer}&\multicolumn{2}{c|}{LogTrans}&\multicolumn{2}{c}{Reformer}\\
			\midrule
			\multicolumn{2}{c|}{Metric} & MSE & MAE& MSE & MAE& MSE & MAE& MSE & MAE& MSE & MAE& MSE & MAE&MSE  & MAE\\
			\midrule
			\multirow{4}{*}{\rotatebox{90}{$ETTm2$}}
			& 96   & \textbf{0.176} & \textbf{0.252} & 0.189 & 0.280& 0.203& 0.287 & 0.255& 0.339& 0.365 & 0.453& 0.768& 0.642& 0.658& 0.619\\
			& 192  & \textbf{0.241} & \textbf{0.296} & 0.253 & 0.319& 0.269& 0.328 & 0.281& 0.340& 0.533 & 0.563& 0.989& 0.757& 1.078& 0.827\\
			& 336  & \textbf{0.303} & \textbf{0.335} & 0.314 & 0.357& 0.325& 0.366 & 0.339& 0.372& 1.363 & 0.887& 1.334& 0.872& 1.549& 0.972\\
			& 720  & \textbf{0.401} & \textbf{0.399} & 0.414 & 0.413& 0.421& 0.415 & 0.422& 0.419& 3.379 &1.388&  3.048& 1.328& 2.631& 1.242\\
			\midrule
			\multirow{4}{*}{\rotatebox{90}{$Electricity$}}
			& 96  &0.204 & \textbf{0.287} & 0.187& 0.304 & \textbf{0.183}& 0.297 & 0.201& 0.317& 0.274& 0.368& 0.258& 0.357 &0.312 &0.402\\
			& 192 &0.208 & \textbf{0.291} & 0.199& 0.315 & \textbf{0.195}& 0.308 & 0.222& 0.334& 0.296& 0.386& 0.266& 0.368 &0.348 &0.433\\
			& 336 &0.222 & \textbf{0.306} & \textbf{0.212}& 0.329& \textbf{0.212}& 0.313 & 0.231& 0.338& 0.300& 0.394& 0.280& 0.380 &0.350 &0.433\\
			& 720 &0.258 & \textbf{0.332} & 0.233& 0.345& \textbf{0.231}& 0.343 & 0.254& 0.361& 0.373& 0.439& 0.283& 0.376 &0.340 &0.420\\
			\midrule
			\multirow{4}{*}{\rotatebox{90}{$Exchange$}}
			& 96  & \textbf{0.082}&  0.206        & 0.085 & \textbf{0.204} & 0.139& 0.276 & 0.197& 0.323& 0.847& 0.752& 0.968& 0.812 &1.065 &0.829\\
			& 192 & 0.184 & 0.314 & \textbf{0.182} & \textbf{0.303} & 0.256& 0.369 & 0.300& 0.369& 1.204& 0.895& 1.040& 0.851 &1.188 &0.906\\
			& 336 & \textbf{0.307}& \textbf{0.416} & 0.348 & 0.428 & 0.426& 0.464 & 0.509& 0.524& 1.672& 1.036& 1.659& 1.081 &1.357 &0.976\\
			& 720 & \textbf{0.554} &\textbf{0.582}  & 1.025 & 0.774 & 1.090& 0.800 & 1.447& 0.941& 2.478& 1.310& 1.941& 1.127 &1.510 &1.016\\
			\midrule
			\multirow{4}{*}{\rotatebox{90}{$Traffic$}} 
			&96   &0.564 &0.350    &0.607  &0.392   & \textbf{0.562}  &\textbf{0.349} &0.613  &0.388  &0.719  &0.391  &0.684  &0.384  &0.732  &0.423 \\
			& 192 &\textbf{0.546} &\textbf{0.345}    &0.621  &0.399 &0.562  &0.346 &0.616  &0.382  &0.696  &0.379  &0.685  &0.390  &0.733  &0.420 \\
			& 336 &\textbf{0.546} &0.340    &0.622  &0.396&0.570  &\textbf{0.323} &0.622  &0.337  &0.777&0.420  &0.733  &0.408  &0.742  &0.420 \\
			& 720 &\textbf{0.565} &\textbf{0.347}    &0.632  &0.396&0.596  &0.368 &0.660  &0.408  &0.864  &0.472 & 0.717 &0.396  &0.755  &0.423 \\
			\midrule
			\multirow{4}{*}{\rotatebox{90}{$Weather$}} 
			& 96  & \textbf{0.185} &\textbf{0.219}  &0.197 &0.281   &0.217  &0.296 &0.266  &0.336  &0.300  &0.384  &0.458  &0.490  &0.689  &0.596 \\
			& 192 & \textbf{0.231} &\textbf{0.256}  &0.237  &0.312   &0.276  &0.336 &0.307 &0.367 &0.598  &0.544  &0.658  &0.589  &0.752 &0.638 \\
			& 336 & \textbf{0.283} &\textbf{0.256}  &0.298 &0.353  &0.339  &0.380 &0.359  &0.395  &0.578 &0.523  &0.797  &0.652  &0.639  &0.596 \\
			& 720 & 0.358&\textbf{0.343}  & \textbf{0.352}  &0.388  &0.403  &0.428 &0.419  &0.428  &1.059  &0.741 & 0.869 &0.675  &1.130  &0.792 \\
			\midrule
			\multirow{4}{*}{\rotatebox{90}{$ILI$}} 
			&24 & \textbf{1.648} & \textbf{0.805}   &2.527  &1.020  &2.203  &0.963 &3.483  &1.287  &5.764  &1.677  &4.480  &1.444  &4.400  &1.382 \\
			&36 & \textbf{1.894} & \textbf{0.837}   &2.615  &1.007  &2.272  &0.976 &3.103  &1.148 &4.755  &1.467  &4.799  &1.467  &4.783  &1.448 \\
			&48 & \textbf{1.962} & \textbf{0.856}   &2.359. &0.972  &2.209  &0.981 &2.669  &1.085  &4.763&1.469  &4.800  &1.468  &4.832  &1.465 \\
			&60 & \textbf{1.970} & \textbf{0.876}   &2.487  &1.016  &2.545  &1.061 &2.770  &1.125  &5.264  &1.564 & 5.278 &1.560  &4.882  &1.483 \\
			\bottomrule
		\end{tabular}
\end{scriptsize}
\label{tab:96multivariate}
\vskip -0.1in
\end{table}

\subsubsection{Parameter Sensitivity }
Here we test the impact of the grid number $J$ and the number of smoothing iterations $\nu_\ell$ on forecasting accuracy. Table \ref{tab:diff-grid} shows the impact of grid number when the number of iterations is equal to 2. We set the number of grids as 2, 3, 4, and 5, and observed that in the trade-off between prediction accuracy and computational efficiency, a grid number of 3 is the best choice.  We then explored the number of iterations within the grid on this condition, setting each grid iteration number as 1, 2, 3, 4, and 5. The results in Table \ref{tab:diff-ite} show that the optimal number of iterations varies for different datasets, the number of iterations equal to 3 is most suitable for the ETTm2, Exchange, and Weather data sets, and the number of iterations equal to 4 is most suitable for the Electricity and Traffic data sets. 

\begin{table}[H]
\centering
\caption{The number of grids under each grid iteration is equal to 2; a lower MSE or MAE indicates a better prediction. The best results are highlighted in bold.}\vspace{-3mm}
\vskip 0.1in
\begin{scriptsize}
		\begin{tabular}{c|c|cccccccc}
			\toprule
			\multicolumn{2}{c|}{Number of grid }&\multicolumn{2}{c|}{2}&\multicolumn{2}{c|}{3}&\multicolumn{2}{c|}{4}&\multicolumn{2}{c}{5}\\
			\midrule
			\multicolumn{2}{c|}{Metric} & MSE & MAE& MSE & MAE& MSE & MAE& MSE & MAE\\
			\midrule
			\multirow{4}{*}{\rotatebox{90}{$ETTm2$}}
			& 96  & 0.244 & 0.317 & 0.183 & 0.268 & 0.176 & 0.266 & \textbf{0.175} & \textbf{0.262}\\
			& 192 & 0.263 & 0.335 & 0.238 & 0.313 & 0.245 & 0.315 & \textbf{0.231} & \textbf{0.298}\\
			& 336 & 0.336 & 0.376 & \textbf{0.283} & \textbf{0.334} & 0.298 & 0.350 & 0.306 & 0.360\\
			& 720 & 0.465 & 0.452 & \textbf{0.378} & \textbf{0.397} & 0.390 & 0.400 & 0.400 & 0.406\\
			\midrule
			\multirow{4}{*}{\rotatebox{90}{$Electricity$}}
			& 96  & 0.243 & 0.376 & 0.154 & 0.262 & \textbf{0.149} & \textbf{0.253} & 0.152 & 0.256\\
			& 192 & 0.253 & 0.367 & \textbf{0.170} & 0.274 & 0.172 & 0.275 & 0.166 & \textbf{0.267}\\
			& 336 & 0.263 & 0.366 & 0.182 & 0.287 & \textbf{0.178} & \textbf{0.280} & 0.201 & 0.301\\
			& 720 & 0.364 & 0.446 & 0.224 & 0.321 & \textbf{0.223} & \textbf{0.318} & 0.226 & 0.322\\
			\midrule
			\multirow{4}{*}{\rotatebox{90}{$Exchange$}}
			& 96  & 0.099 & 0.230 & 0.105 & 0.234 & \textbf{0.086} & \textbf{0.216} & 0.105 & 0.236\\
			& 192 & 0.312 & 0.423 & 0.273 & 0.384 & \textbf{0.228} & \textbf{0.353} & 0.196 & 0.326\\
			& 336 & 0.336 & 0.446 & 0.337 & 0.434 & \textbf{0.319} & \textbf{0.428} & 0.332 & 0.435\\
			& 720 & 1.405 & 0.924 & \textbf{0.530} & \textbf{0.562} & 1.026 & 0.764 & 0.980 & 0.751\\
			\midrule
			\multirow{4}{*}{\rotatebox{90}{$Traffic$}} 
			& 96  & 0.524 & 0.399 & \textbf{0.419} & \textbf{0.293} & 0.421 & 0.297 & 0.419 & 0.296\\
			& 192 & 0.545 & 0.309 & \textbf{0.431} & \textbf{0.303} & 0.441 & 0.306 & 0.432 & 0.299\\
			& 336 & 0.550 & 0.414 & \textbf{0.446} & \textbf{0.305} & 0.448 & 0.307 & 0.447 & 0.306\\
			& 720 & 0.578 & 0.424 & \textbf{0.470} & \textbf{0.311} & 0.477 & 0.318 & 0.471 & 0.322\\
			\midrule
			\multirow{4}{*}{\rotatebox{90}{$Weather$}} 
			& 96  & 0.180 & 0.265 & \textbf{0.158} & 0.221 & 0.159 & 0.209 & 0.160 & 0.209\\
			& 192 & 0.259 & 0.345 & 0.205 & 0.251 & \textbf{0.200} & \textbf{0.246} & 0.203 & 0.251\\
			& 336 & 0.329 & 0.326 & \textbf{0.242} & \textbf{0.280} & 0.246 & 0.281 & 0.247 & 0.292\\
			& 720 & 0.374 & 0.401 & \textbf{0.310} & \textbf{0.333} & 0.315 & 0.335 & 0.317 & 0.342\\
			\bottomrule
		\end{tabular}
\end{scriptsize}
\label{tab:diff-grid}
\vskip -0.1in
\end{table}

\begin{table}[H]
\centering
\caption{The number of iterations under the grid number is 3; a lower MSE or MAE indicates a better prediction. The best results are highlighted in bold.}\vspace{-3mm}
\vskip 0.1in
\begin{scriptsize}
	
		\begin{tabular}{c|c|cccccccccc}
			\toprule
			\multicolumn{2}{c|}{Number of iteration }&\multicolumn{2}{c|}{1}&\multicolumn{2}{c|}{2}&\multicolumn{2}{c|}{3}&\multicolumn{2}{c|}{4}&\multicolumn{2}{c}{5}\\
			\midrule
			\multicolumn{2}{c|}{Metric} & MSE & MAE & MSE & MAE& MSE & MAE& MSE & MAE& MSE & MAE\\
			\midrule
			\multirow{4}{*}{\rotatebox{90}{$ETTm2$}}
			& 96  & 0.176 & 0.268 & 0.183 & 0.268 & \textbf{0.173} & \textbf{0.253} & 0.174 & 0.261 & 0.175 & 0.265\\
			& 192 & 0.239 & 0.313 & 0.238 & 0.313 & \textbf{0.230} & \textbf{0.296} & 0.230 & 0.300 & 0.233 & 0.301\\
			& 336 & 0.281 & 0.334 & 0.283 & 0.334 & \textbf{0.279} & \textbf{0.329} & 0.280 & 0.329 & 0.282 & 0.330\\
			& 720 & 0.386 & 0.415 & 0.378 & 0.415 & \textbf{0.367} & \textbf{0.385} & 0.377 & 0.393 & 0.379 & 0.395\\
			\midrule
			\multirow{4}{*}{\rotatebox{90}{$Electricity$}}
			& 96  & 0.145 & \textbf{0.248} & 0.154 & 0.262 & 0.161 & 0.266 & \textbf{0.144} & 0.250 & 0.202 & 0.301\\
			& 192 & 0.163 & 0.265 & 0.170 & 0.274 & 0.166 & 0.267 & \textbf{0.163} & \textbf{0.262} & 0.178 & 0.279\\
			& 336 & 0.180 & 0.281 & 0.182 & 0.287 & 0.182 & 0.284 & \textbf{0.176} & \textbf{0.276} & 0.186 & 0.287\\
			& 720 & 0.221 & 0.318 & 0.224 & 0.321 & 0.217 & 0.314 & \textbf{0.212} & \textbf{0.308} & 0.216 & 0.314\\
			\midrule
			\multirow{4}{*}{\rotatebox{90}{$Exchange$}}
			& 96  & 0.084 & 0.211 & 0.105 & 0.234 & \textbf{0.082} & \textbf{0.206} & 0.087 & 0.216 & 0.083 & 0.208\\
			& 192 & 0.161 & 0.295 & 0.273 & 0.384 & \textbf{0.184} & \textbf{0.314} & 0.208 & 0.338 & 0.186 & 0.315\\
			& 336 & 0.336 & 0.433 & 0.337 & 0.434 & \textbf{0.307} & \textbf{0.416} & 0.314 & 0.427 & 0.327 & 0.431\\
			& 720 & 0.996 & 0.757 & \textbf{0.530} & \textbf{0.562} & 0.554 & 0.582 & 0.632 & 0.613 & 0.916 & 0.719\\
			\midrule
			\multirow{4}{*}{\rotatebox{90}{$Traffic$}} 
			& 96  & 0.413 & 0.289 & 0.419 & 0.293 & 0.423 & 0.296 & \textbf{0.396} & \textbf{0.285} & 0.429 & 0.299\\
			& 192 & 0.434 & 0.301 & 0.436 & 0.303 & 0.435 & 0.301 & \textbf{0.417} & \textbf{0.295} & 0.440 & 0.304\\
			& 336 & 0.448 & 0.303 & 0.446 & 0.305 & 0.445 & 0.304 & \textbf{0.436} & \textbf{0.302} & 0.448 & 0.305\\
			& 720 & 0.472 & 0.314 & 0.472 & 0.318 & 0.470 & \textbf{0.313} & \textbf{0.468} & 0.315 & 0.472 & 0.315\\
			\midrule
			\multirow{4}{*}{\rotatebox{90}{$Weather$}} 
			& 96  & 0.156 & 0.206 & 0.165 & 0.221 & \textbf{0.155} & \textbf{0.196} & 0.160 & 0.217 & 0.174 & 0.265\\
			& 192 & \textbf{0.199} & 0.246 & 0.205 & 0.251 & 0.201 & \textbf{0.239} & 0.200 & 0.251 & 0.232 & 0.300\\
			& 336 & 0.247 & 0.283 & 0.248 & 0.284 & \textbf{0.244} & \textbf{0.279} & 0.244 & 0.284 & 0.280 & 0.329\\
			& 720 & 0.321 & 0.351 & 0.310 & 0.333 & 0.313 & \textbf{0.329} & \textbf{0.309} & 0.332 & 0.367 & 0.385\\
			\bottomrule
		\end{tabular}
\end{scriptsize}
\label{tab:diff-ite}
\vskip -0.1in
\end{table}

\subsubsection{Input length}
We considered the influence of different input lengths on the forecasting error by setting the input length $I \in \{24, 36,48 ,60\}$ for the ILI data set and $I \in \{96, 192, 336, 720\}$ for the other datasets. The results are shown in Table \ref{tab:diff-input-length}. With the exception of the Exchange dataset, the forecasting error can significantly decrease as the input length increases. This shows that with the increase in length, FV-MgNET can better extract relevant features in the time series and make more accurate forecasting.
\begin{table}[htbp]
\centering
\caption{Multivariate long sequence time-series forecasting results on different input lengths. ILI input length $I \in {24, 36,48 ,60}$ and others $I \in {96, 192, 336, 720}$ }\vspace{-3mm}
\vskip 0.1in
\small
\begin{scriptsize}
		\begin{tabular}{c|c|cccccccc}
			\toprule
			\multicolumn{2}{c|}{Input length }&\multicolumn{2}{c|}{96}&\multicolumn{2}{c|}{192}&\multicolumn{2}{c|}{336}&\multicolumn{2}{c}{720}\\
			\midrule
			\multicolumn{2}{c|}{Metric} & MSE & MAE& MSE & MAE& MSE & MAE& MSE & MAE\\
			\midrule
			\multirow{4}{*}{\rotatebox{90}{$ETTm2$}}
			& 96  & 0.176 & 0.252 & 0.173 & 0.251 & \textbf{0.173} & \textbf{0.253} & 0.180 & 0.260\\
			& 192 & 0.241 & 0.296 & 0.234 & \textbf{0.291} & \textbf{0.230} & 0.296 & 0.249 & 0.303\\
			& 336 & 0.303 & 0.335 & 0.295 & 0.330 & \textbf{0.279} & \textbf{0.329} & 0.304 & 0.341\\
			& 720 & 0.401 & 0.393 & 0.386 & 0.387 & \textbf{0.367} & \textbf{0.385} & 0.398 & 0.397\\
			\midrule
			\multirow{4}{*}{\rotatebox{90}{$Electricity$}}
			& 96  & 0.204 & 0.287 & 0.162 & 0.259 & \textbf{0.144} & \textbf{0.250} & 0.151 & 0.252\\
			& 192 & 0.208 & 0.291 & 0.179 & 0.274 & \textbf{0.163} & \textbf{0.262} & 0.167 & 0.267\\
			& 336 & 0.222 & 0.306 & 0.194 & 0.289 & \textbf{0.176} & \textbf{0.276} & 0.186 & 0.286\\
			& 720 & 0.258 & 0.332 & 0.231 & 0.316 & \textbf{0.212} & \textbf{0.308} & 0.229 & 0.319\\
			\midrule
			\multirow{4}{*}{\rotatebox{90}{$Exchange$}}
			& 96  & \textbf{0.082} & \textbf{0.206} & 0.101 & 0.225 & 0.119 & 0.242 & 0.110 & 0.240\\
			& 192 & \textbf{0.184} & \textbf{0.314} & 0.253 & 0.353 & 0.251 & 0.352 & 0.245 & 0.366\\
			& 336 & \textbf{0.307} & \textbf{0.416} & 0.482 & 0.495 & 0.435 & 0.473 & 0.482 & 0.503\\
			& 720 & \textbf{0.554} & \textbf{0.582} & 0.948 & 0.726 & 1.219 & 0.778 & 1.370 & 0.830\\
			\midrule
			\multirow{4}{*}{\rotatebox{90}{$Traffic$}} 
			& 96  & 0.564 & 0.350 & 0.446 & 0.302 & 0.414 & 0.290 & \textbf{0.396} & \textbf{0.285}\\
			& 192 & 0.546 & 0.345 & 0.463 & 0.312 & 0.435 & 0.300 & \textbf{0.417} & \textbf{0.295}\\
			& 336 & 0.546 & 0.340 & 0.476 & 0.314 & 0.449 & 0.304 & \textbf{0.436} & \textbf{0.302}\\
			& 720 & 0.565 & 0.347 & 0.494 & 0.323 & 0.473 & 0.315 & \textbf{0.468} & \textbf{0.315}\\
			\midrule
			\multirow{4}{*}{\rotatebox{90}{$Weather$}} 
			& 96  & 0.185 & 0.219 & 0.167 & 0.203 & \textbf{0.155} & \textbf{0.196} & 0.153 & 0.198\\
			& 192 & 0.231 & 0.256 & 0.213 & 0.244 & \textbf{0.201} & \textbf{0.239} & 0.203 & 0.245\\
			& 336 & 0.283 & 0.296 & 0.263 & 0.284 & \textbf{0.244} & \textbf{0.279} & 0.256 & 0.285\\
			& 720 & 0.358 & 0.343 & 0.340 & 0.337 & \textbf{0.313} & \textbf{0.329} & 0.330 & 0.335\\
			\midrule
			\multicolumn{2}{c|}{Input length }&\multicolumn{2}{c|}{24}&\multicolumn{2}{c|}{36}&\multicolumn{2}{c|}{48}&\multicolumn{2}{c}{60}\\
			\midrule
			\multirow{4}{*}{\rotatebox{90}{$ILI$}} 
			& 24  & 3.487 & 1.067 & 1.648 & 0.805 & 1.787 & 0.792 & \textbf{1.647} & \textbf{0.764}\\
			& 36  & 3.785 & 1.135 & 1.894 & 0.837 & 1.990 & 0.873 & \textbf{1.841} & \textbf{0.839}\\
			& 48  & 2.487 & 0.987 & 1.962 & 0.856 & 1.845 & 0.835 & \textbf{1.831} & \textbf{0.853}\\
			& 60  & 2.387 & 0.971 & 1.970 & 0.876 & 1.812 & 0.824 & \textbf{1.765} & \textbf{0.814}\\
			\bottomrule
		\end{tabular}
\end{scriptsize}
\label{tab:diff-input-length}
\vskip -0.1in
\end{table}

\subsubsection{ETT full benchmark}
We tested our model in the four ETT multivariate data sets and the results are shown in Table \ref{tab:multi-benchmarks-ett}. ETTh and ETTm have different recording frequencies; the ETTh1 and ETTh2 data sets are recorded hourly, whereas ETTm1 and ETTm2 are recorded every 15 minutes. We obtained comparable results with extensive baseline models and, especially considering the MAE error in the ETTh1 and ETTh2 data sets, FV-MgNet achieves state-of-the-art results.
\begin{table*}[htbp]
\centering
\caption{Multivariate long sequence time-series forecasting results on ETT full benchmark. The best results are highlighted in bold.}\vspace{-1mm}
\begin{small}
	\scalebox{0.79}{
		\begin{tabular}{c|c|cccccccccccccccc}
			\toprule
			\multicolumn{2}{c|}{Methods}&\multicolumn{2}{c|}{FV-MgNet}&\multicolumn{2}{c|}{FiLM}&\multicolumn{2}{c|}{FEDformer}&\multicolumn{2}{c|}{Autoformer}&\multicolumn{2}{c|}{Informer}&\multicolumn{2}{c|}{LogTrans}&\multicolumn{2}{c}{Reformer}\\
			\midrule
			\multicolumn{2}{c|}{Metric} & MSE & MAE& MSE  & MAE& MSE & MAE& MSE  & MAE& MSE  & MAE& MSE  & MAE&MSE  & MAE\\
			\midrule
			\multirow{4}{*}{\rotatebox{90}{$ETTh1$}}
			& 96  &\textbf{0.364} &\textbf{0.382} & 0.371& 0.394& 0.376 & 0.419 & 0.449 & 0.459 & 0.865 & 0.713& 0.878& 0.740& 0.837& 0.728\\
			& 192 &\textbf{0.409} &\textbf{0.415} & 0.414& 0.423& 0.420 & 0.448 & 0.500 & 0.482 & 1.008 & 0.792& 1.037& 0.824& 0.923& 0.766\\
			& 336 &0.452 &\textbf{0.443} & \textbf{0.442}& 0.445& 0.459 & 0.465 & 0.521 & 0.496 & 1.107 & 0.809& 1.238& 0.932& 1.097& 0.835\\
			& 720 &0.473 &0.482 & \textbf{0.465}& \textbf{0.472}& 0.506 & 0.507 & 0.514 & 0.512 & 1.181 & 0.865&  1.135& 0.852& 1.257& 0.889\\
			\midrule
			\multirow{4}{*}{\rotatebox{90}{$ETTh2$}}
			& 96  &\textbf{0.278} &\textbf{0.382} & 0.284& 0.397& 0.346 & 0.388 & 0.358 & 0.397 & 3.755& 1.525& 2.116& 1.197 &2.626 &1.317\\
			& 192 &0.360 &\textbf{0.436} & \textbf{0.357}& 0.452& 0.429 & 0.439 & 0.456 & 0.452 & 5.602& 1.931& 4.315& 1.635 &11.12 &2.979\\
			& 336 &\textbf{0.365} &\textbf{0.457} & 0.377& 0.486& 0.482 & 0.480 & 0.482 & 0.486 & 4.721& 1.835& 1.124& 1.604 &9.323 &2.769\\
			& 720 &\textbf{0.418} &\textbf{0.432} & 0.439& 0.456& 0.463 & 0.474 & 0.515 & 0.511 & 3.647& 1.625& 3.188& 1.540 &3.874 &1.697\\
			\midrule
			\multirow{4}{*}{\rotatebox{90}{$ETTm1$}}
			& 96  & 0.320 & 0.352 &\textbf{0.302} &\textbf{0.345}& 0.378 & 0.418 & 0.505 & 0.475 & 0.672& 0.571& 0.600& 0.546 &0.538 &0.528\\
			& 192 & 0.363 & 0.380 &\textbf{0.338} &\textbf{0.368}& 0.426 & 0.441 & 0.553 & 0.496 & 0.795& 0.669& 0.837& 0.700 &0.658 &0.592\\
			& 336 & 0.400 & 0.407 &\textbf{0.373} &\textbf{0.388}& 0.445 & 0.459 & 0.621 & 0.537 & 1.212& 0.871& 1.124& 0.832 &0.898 &0.721\\
			& 720 & 0.456 & 0.441 &\textbf{0.420} &\textbf{0.420}& 0.543 & 0.490 & 0.671 & 0.561 & 1.166& 0.823& 1.153& 0.820 &1.102 &0.841\\
			\midrule
			\multirow{4}{*}{\rotatebox{90}{$ETTm2$}} 
			&96  & 0.173 & 0.253 &\textbf{0.165} &\textbf{0.256}          &0.203 & 0.287 & 0.255 & 0.339 &0.365  &0.453  &0.768  &0.642  &0.658  &0.619 \\
			&192 & 0.240 & \textbf{0.296} &\textbf{0.222} &\textbf{0.296} &0.269 & 0.328 & 0.281 & 0.340 &0.533  &0.563  &0.989  &0.757  &1.078  &0.827 \\
			&336 & 0.296 & \textbf{0.333} &\textbf{0.277} &\textbf{0.333} &0.325 & 0.366 & 0.339 & 0.372 &1.363  &0.887  &1.334  &0.872  &1.549  &0.972 \\
			&720 & 0.378 & \textbf{0.385} &\textbf{0.371} &0.389          &0.421 & 0.415 & 0.422 & 0.419 &1.338  &3.379  & 3.048 &1.328  &2.631  &1.242 \\
			\bottomrule
		\end{tabular}
	}
\end{small}
\label{tab:multi-benchmarks-ett}
\vskip -0.1in
\end{table*}

\section{Conclutions}\label{conclution}

The iterative method and multigrid method in numerical PDEs have well-developed mathematical theories. By considering the forecasting problems from the constrained model point of view, and taking advantage of the multigrid method and MgNet, we propose FV-MgNet models for long-term time series forecasting. We investigated variants of FV-MgNet and found that the use of fully connected MgNet with V-cycle gives the best results. In comparison with state-of-the-art models, we demonstrate the good performance of FV-MgNet on different data sets, which achieves better results with fewer parameters and faster inference speed. Furthermore, the results obtained by FV-MgNet significantly improve as the length of the input sequence increases.

\bibliographystyle{elsarticle-num}
\bibliography{fvmgnet.bib}






\end{document}